\documentclass[letterpaper]{article} 
\usepackage{aaai2026}  
\usepackage{times}  
\usepackage{helvet}  
\usepackage{courier}  
\usepackage[hyphens]{url}  
\usepackage{graphicx} 
\usepackage{amssymb}
\usepackage{amsmath}
\usepackage[linesnumbered,ruled,vlined]{algorithm2e}
\usepackage{algpseudocode}
\usepackage{booktabs}
\usepackage[dvipsnames,table,xcdraw]{xcolor}
\usepackage{multirow}
\usepackage{diagbox}
\usepackage{natbib}  
\usepackage{caption} 
\usepackage{bm}

\usepackage{newfloat}
\usepackage{listings}
\DeclareCaptionStyle{ruled}{labelfont=normalfont,labelsep=colon,strut=off} 
\usepackage{newfloat}
\usepackage{listings}

\title{Unlocking Robust Semantic Segmentation Performance\\ via Label-only Elastic Deformations against Implicit Label Noise}
\author{
    Yechan Kim\textsuperscript{\rm 1}\thanks{Corresponding authors: Moongu Jeon; Yechan Kim.}\equalcontrib, Dongho Yoon\textsuperscript{\rm 1}\equalcontrib, Younkwan Lee\textsuperscript{\rm 2}, Unse Fatima\textsuperscript{\rm 1}, Hong Kook Kim\textsuperscript{\rm 1},\\Songjae Lee\textsuperscript{\rm 3}, Sanga Park\textsuperscript{\rm 3}, Jeong Ho Park\textsuperscript{\rm 3}, Seonjong Kang\textsuperscript{\rm 3}, Moongu Jeon\textsuperscript{\rm 1}\\
}
\affiliations{
    \textsuperscript{\rm 1}GIST, \textsuperscript{\rm 2}Samsung Electronics, \textsuperscript{\rm 3}LIG Nex1\\

    \textsuperscript{\rm 1}\{yechankim, gidong76, unse.fatima\}@gm.gist.ac.kr, \{hongkook, mgjeon\}@gist.ac.kr, \\
    \textsuperscript{\rm 2}youn720.lee@samsung.com, \textsuperscript{\rm 3}\{songjae.lee, sanga.park, jeongho.park1, seonjong.kang\}@lignex1.com
}

\usepackage{bibentry}

\begin{document}

\maketitle
\begin{abstract}
While previous studies on image segmentation focus on handling severe (or explicit) label noise, real-world datasets also exhibit subtle (or implicit) label imperfections.
    These arise from inherent challenges, such as ambiguous object boundaries and annotator variability.
Although not explicitly present, such mild and latent noise can still impair model performance.
    Typical data augmentation methods, which apply identical transformations to the image and its label, risk amplifying these subtle imperfections and limiting the model's generalization capacity.
In this paper, we introduce \textbf{\textit{NSegment+}}, a novel augmentation framework that decouples image and label transformations to address such realistic noise for semantic segmentation.
    By introducing controlled elastic deformations only to segmentation labels while preserving the original images, our method encourages models to focus on learning robust representations of object structures despite minor label inconsistencies.
Extensive experiments demonstrate that \textbf{\textit{NSegment+}} consistently improves performance, achieving mIoU gains of up to +2.29, +2.38, +1.75, and +3.39 in average on Vaihingen, LoveDA, Cityscapes, and PASCAL VOC, respectively—even without bells and whistles, highlighting the importance of addressing implicit label noise.  
    These gains can be further amplified when combined with other training tricks, including CutMix and Label Smoothing.

\end{abstract}

\vspace{-0.4cm}
\begin{links}
    \link{Code}{https://github.com/unique-chan/NSegmentPlus}
\end{links}

\section{Introduction}

\begin{figure}[!t]
    \centering
    \includegraphics[width=8.6cm]{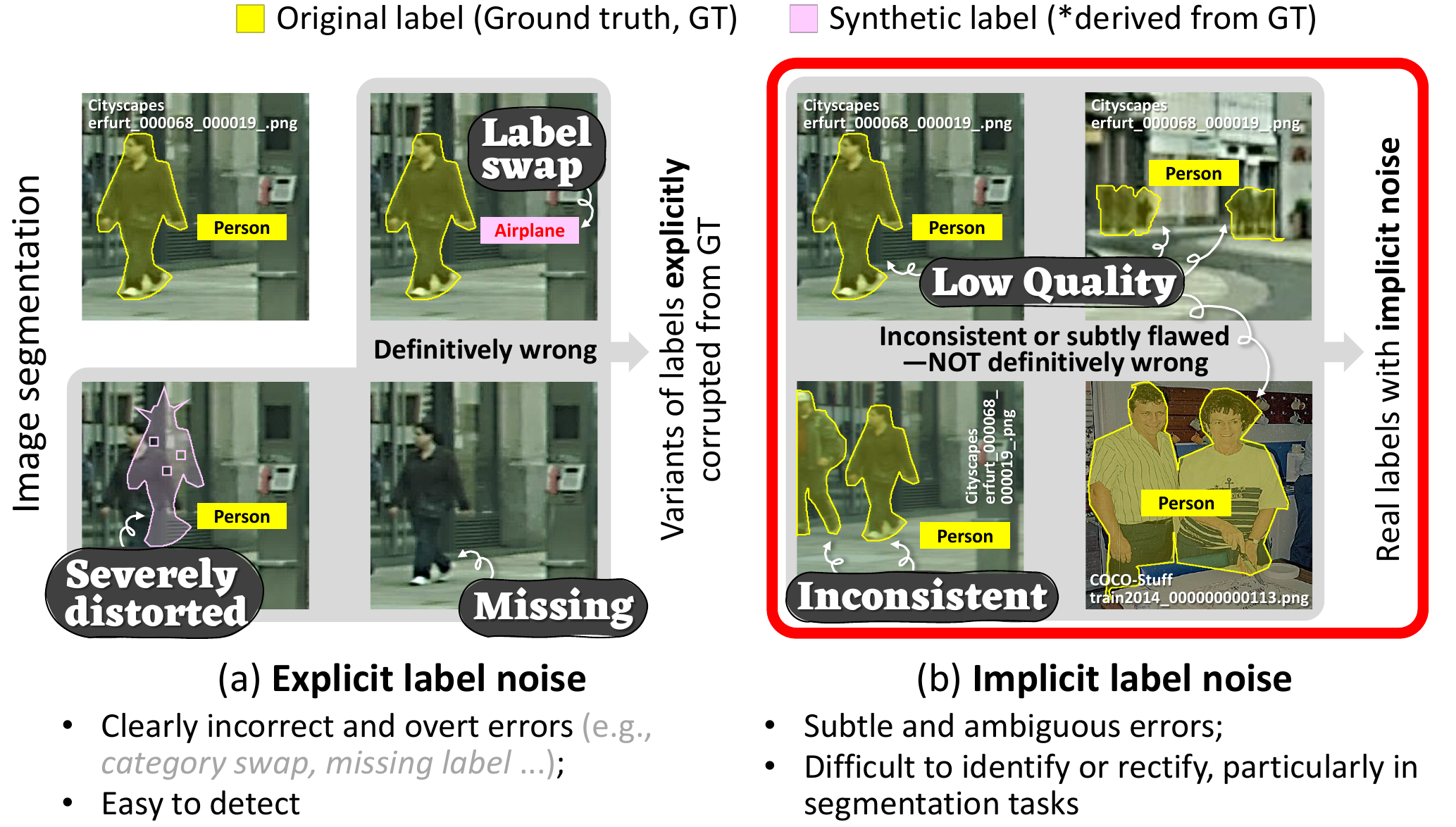} \vspace{-0.5cm}
    \caption{Illustration of (a) explicit and (b) implicit label noise in semantic segmentation.
    The red box (\textbf{\textcolor{red}{$\Box$}}) indicates the core challenge targeted in this work.}
    \label{fig1}
    \vspace{-0.5cm}
\end{figure}

Semantic segmentation, a fundamental task in computer vision, involves assigning semantic labels to each pixel of an image. 
    Despite significant progress in model architectures, the performance of segmentation models remains heavily reliant on the quality of the annotated datasets \cite{brar2025image}.
However, creating high-quality pixel-level annotations is not only costly and time-consuming but also prone to subtle imperfections.
    Even meticulously curated datasets often contain hidden label noise (which we refer to as `\textit{implicit}' label noise) due to inherent challenges, such as ambiguous object boundaries, mixed pixels, shadows, occlusions, and inconsistencies among annotators \cite{wang2021loveda}. 
Unlike severe and overt label noise (hereinafter referred to as `\textit{explicit}' label noise), including missing masks and corrupted categories, these implicit imperfections are relatively hard to detect and rectify even for annotation experts, as shown in Fig. \ref{fig1}.
       Yet, even if individual instances of subtle label noise seem harmless, their inconsistent distribution can collectively degrade model performance.

Recent advances in Learning from Noisy Labels (LNL) have introduced a variety of strategies, including model architecture modification, label refinement mechanism, and loss function design, to combat the effects of inaccurate supervision \cite{shen2023survey}.
    While these methods have shown promise in mitigating explicit label errors, they often rely on complex training pipelines and demand extensive hyperparameter tuning.
More critically, our experimental results reveal that existing LNL methods are suboptimal when confronted with implicit label noise, which is often overlooked in prior work.
    Such implicit noise, though inconspicuous, can erode model performance.
To mitigate this issue, we seek an alternate paradigm: a lightweight but efficient solution for handling implicit annotation noise in existing image segmentation benchmarks.

A practical alternative lies in perturbing the training data itself—namely, through data augmentation.
    By exposing the model to multiple augmented variants of the same input, it can learn to focus on generalizable patterns rather than memorizing specific noise. 
        Typical augmentation strategies \cite{islam2024diffusemix}
        apply identical geometric transformations to both the image and its corresponding label.
    The problem is that such synchronized augmentations can inadvertently amplify annotation noise, preserving or even intensifying structural inconsistencies in the labels.

To address this challenge, we propose \textbf{\textit{NSegment+}}, a novel data augmentation framework that decouples transformations applied to images and labels.
    Unlike conventional augmentation, which treats labels as perfect ground truth, our method acknowledges the presence of hidden label uncertainty and leverages it to improve model robustness.
        By keeping the input image unchanged and only distorting the segmentation masks, our method encourages the model to rely less on possibly noisy label boundaries and instead focus on robust semantic cues inherent in the image.
    For label-specific deformations, we introduce three key innovations to enhance deformation stability and scalability: \vspace{-0.1cm}
\begin{itemize}
    \item We are the first to \textbf{generalize the use of elastic deformation} beyond its prevalent application in medical image segmentation \cite{islam2024systematic}, and show its effectiveness for general semantic segmentation tasks. \vspace{-0.1cm} 
    \item We incorporate \textbf{per-sample, per-epoch stochastic deformation}, where each segmentation mask is independently augmented at every training epoch using a randomly sampled combination of `deformation magnitude' and `spatial smoothness'. This simple yet powerful mechanism injects high variability into the training process, serving as a form of label-level regularization. \vspace{-0.1cm}
    \item We design a \textbf{scale-aware deformation suppression} mechanism to protect small objects from excessive distortion. During the augmentation process, deformation fields are selectively masked around small label regions.
        This component is vital as aggressive deformation for small masks may otherwise lead to semantic erosion.
\end{itemize}

To the best of our knowledge, we first define and address implicit label noise for semantic segmentation.
    To validate the generalizability of \textbf{\textit{NSegment+}}, we conduct extensive experiments on six diverse benchmarks—spanning remote sensing (ISPRS Vaihingen, Potsdam, LoveDA) and natural scenes (Pascal VOC 2012, Cityscapes, COCO-Stuff 10K)—using numerous state-of-the-art semantic segmentation models.
    Overall, \textbf{\textit{NSegment+}} demonstrates significant performance gains, validating its effectiveness. 
   We also demonstrate that \textbf{\textit{NSegment+}} can synergize effectively with other augmentation and regularization schemes, leading to further performance improvements. 

\begin{figure*}[!ht]
    \centering
    \includegraphics[width=15.5cm]{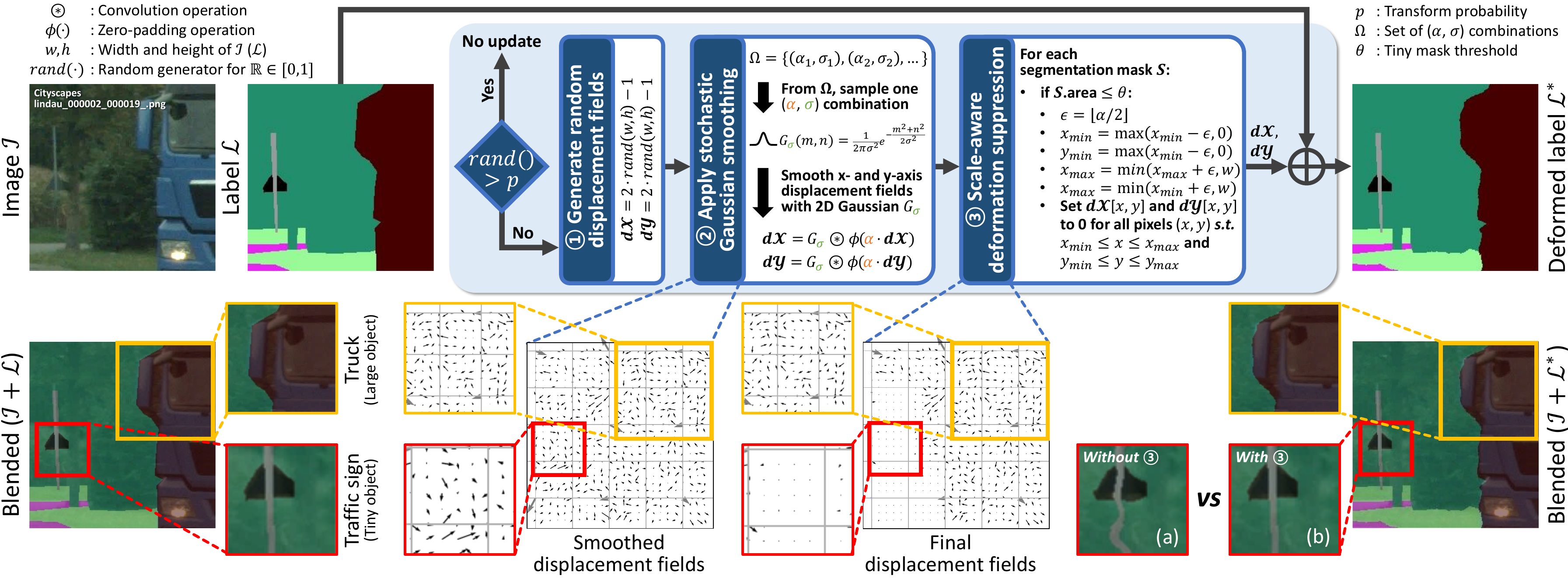} \vspace{-0.3cm}
    \caption{Overall pipeline of the proposed \textbf{\textit{NSegment+}} for data augmentation suited to semantic segmentation. 
        To alleviate implicit label noise, our method perturbs only the label mask while keeping the input image unchanged. 
    Specifically, we introduce two key components: (1) stochastic Gaussian smoothing to diversify labels, and (2) a constraint that suppresses distortions in small masks to retain the structural integrity of labels.
        Especially, the latter matters in our pipeline to avoid severe semantic misalignment as indicated in (a), compared to (b).
    Best viewed on a colored screen with zoom.}
    \label{fig2}
    \vspace{-0.3cm}
\end{figure*}

\section{Related Work}

\subsection{Elastic deformation in computer vision}

Elastic deformation refers to a class of smooth, non-rigid transformations that alter the shape of an image while preserving its continuity and structural integrity.
    Originally grounded in material science \cite{rivlin1948large}, this concept has been widely adopted in computer vision to improve model robustness and generalization.

For instance, \cite{simard2003best} were among the first to apply random elastic distortions to training data in handwritten character recognition.
     In the field of image registration and shape analysis, Large Deformation Diffeomorphic Metric Mapping (LDDMM) \cite{glaunes2008large} is a notable framework based on elastic transformation principles.
Similarly, DeepFlow \cite{weinzaepfel2013deepflow} incorporates elastic-like warping to refine motion estimation.
    Recently, elastic deformation has seen growing application in medical image analysis, to introduce plausible anatomical variability during training, thus increasing the invariance of models to structural differences across patients \cite{islam2024systematic}.

Nevertheless, elastic deformation has been seldom adopted in general image segmentation tasks (e.g., earth observation, urban/driving scene understanding, etc.).
    This hesitation mainly stems from the fact that naive non-rigid warping tends to introduce textual artifacts into the input image, thereby injecting unrealistic visual cues that hinder effective model generalization.
        For instance, unlike CT images, RGB imagery exhibits significantly higher visual variability, hence it is vulnerable to texture misrepresentation.
To bridge this gap, we propose a dedicated augmentation pipeline that enables elastic deformation to be safely and effectively integrated into generic image segmentation.

\subsection{Learning from noisy labels in computer vision}
Handling label noise has been a critical challenge in various computer vision tasks \cite{natarajan2013learning, song2022learning}.
    Over the past decade, the problem of learning with noisy labels (LNL) has received significant attention, especially in image classification.
Existing approaches can largely be categorized into three main directions: robust architecture construction, label cleansing, and robust loss function design.    
     Early efforts focused on modifying the model architecture to handle noisy supervision. 
        These introduced mechanisms, such as noise adaptation layers, to absorb the effects of corrupted labels \cite{sukhbaatar2014training, xiao2015learning, goldberger2017training}. 

Subsequent research turned toward label cleansing strategies, including sample selection \cite{tanaka2018joint, pleiss2020identifying, huang2023twin} or label correction \cite{yu2019does, li2020error, sumbul2023generative}. 
    These methods often require additional modules such as teacher models or auxiliary encoder/decoders \cite{kim2024learning}.
In parallel, a substantial line of work has focused on designing robust loss functions that tolerate label noise without requiring explicit noise correction.
    Examples include bootstrapping strategies \cite{reed2015training, zhou2024l2b} and regularization techniques that prevent memorization of incorrect labels \cite{liu2020early, kang2021noise, aksoy2024multi}.

Meanwhile, a growing body of work has begun to extend LNL techniques from classification to segmentation, mainly by incorporating label refinement mechanisms into the segmentation pipeline \cite{ibrahim2020semi, zheng2021rectifying, liu2022adaptive, kaiser2023compensation, landgraf2024uncertainty, liu2024aio2}.

Despite these advances, most existing studies rely heavily on synthetic noise settings (See Fig. \ref{fig1}(a)), which—although controlled and analytically convenient—fail to reflect the complex and unstructured nature of real-world noise distributions \cite{wei2021learning, chen2021noise}.
    Moreover, current literature predominantly targets explicit noise, where label corruption is easily detectable. 
However, in real datasets, label noise is often implicit, arising from ambiguous object boundaries, occlusions, or annotator variability, as shown in Fig. \ref{fig1}(b), which these methods do not effectively address.

\subsection{Decoupled image-label transformations}
Most LNL studies assume the presence of both category and localization noise.
    However, only a few works have specifically addressed noisy localization as an independent challenge, though model training—particularly for spatial prediction tasks—is often highly sensitive to even small localization errors.
In object detection, \cite{liu2022robust} pioneered this research line by explicitly modeling bounding box (BBox) noise using three types of linear transformations (scaling, rotation, translation), and proposed localization-aware loss objectives aimed at correcting such noisy annotations. 
Rather than attempting to clean or recover the true labels, \cite{kim2025nbbox} focuses on improving robustness by exposing the model to diverse variants of BBox annotations.
    It introduces various perturbations solely to the localization labels, while keeping the corresponding image unchanged, thereby decoupling image and label transformations.
We extend this research direction to semantic segmentation, where implicit localization label noise is further spatially diffuse across the pixel space in existing benchmarks.

\vspace{-0.0cm}

\section{Methodology}
Real-world semantic segmentation datasets often suffer from implicit label noise.
    To overcome this limitation, we design a lightweight yet effective augmentation mechanism, named \textbf{\textit{NSegment+}}, that operates exclusively on segmentation labels to simulate real-world annotation ambiguity.

To facilitate a clearer understanding of our proposed method, we first introduce a basic form of label-level deformation, which we refer to as \textbf{\textit{NSegment}}.
    We then present our final framework, \textbf{\textit{NSegment+}}, which further improves robustness by incorporating scale-aware perturbation control over label-specific deformations.

\subsection{\textbf{\textit{NSegment}}: Label-specific deformation-based data augmentation for semantic segmentation}
This subsection introduces our core augmentation scheme, \textbf{\textit{NSegment}}, which applies elastic deformations only to segmentation labels.
    The transformation is spatially smooth and dynamically varied across training epochs to simulate soft boundary uncertainty and implicit label noise.

Given an image $\mathcal{I} \in \mathbb{R}^{w \times h \times 3}$ and its corresponding segmentation label $\mathcal{L}=\{S_{j} \}_{j=1}^{C}$, where $S_{j} \in \mathbb{R}^{w \times h}$ denotes the binary mask for class $j$, our pipeline proceeds as: \vspace{-0.1cm} 
\begin{enumerate}
    \item \textbf{Generation of random displacement fields}: To introduce local non-rigid perturbations, we first generate two displacement fields $d\mathcal{X}$ and $d\mathcal{Y}$ of shape $w \times h$, as shown in Fig. \ref{fig2}-\textcircled{1}. 
        Each value is drawn from a uniform distribution in $[-1, 1]$.  
    \vspace{-0.1cm}
    \item \textbf{Stochastic smoothing with Gaussian kernel}: 
        As illustrated in Fig. \ref{fig2}-\textcircled{2}, a random pair $(\alpha, \sigma)$ is drawn from a predefined set $\Omega$, where $\alpha$ controls the magnitude (strength) of deformation and $\sigma$ controls the spatial smoothness (via a 2D Gaussian kernel).
            Then, each value in the raw displacement fields $d\mathcal{X}$ and $d\mathcal{Y}$ is scaled by $\alpha$ and $\phi(\cdot)$ applies zero-padding to maintain the original spatial shape.
        Finally, the displacement fields $d\mathcal{X}$ and $d\mathcal{Y}$ are convolved with the Gaussian kernel ${G}_{\sigma}$.
            This operation produces smooth and spatially coherent deformations, enabling label perturbations that mimic soft and realistic implicit annotation noise without introducing sharp discontinuities.
    \vspace{-0.1cm}
    \item \textbf{Elastic warping of segmentation mask label}: For each class-wise label ${S}_{j}$, the $x$- and $y$-axis smoothed displacement fields $d\mathcal{X}$ and $d\mathcal{Y}$ are applied to remap pixel coordinates.
         Specifically, each pixel $(x, y)$ in ${S}_{j}$ is shifted to $(x+d\mathcal{X}[x, y], y+d\mathcal{Y}[x, y])$, using bilinear interpolation with proper boundary clipping ($\varphi(\cdot)$ in Algorithm 1).
     This yields the deformed label ${S}^{*}_{j}$ for each class $j$.

    \vspace{-0.1cm}
\end{enumerate}

Note that the above procedure is applied independently to each image-label pair at every training epoch and is activated only when $rand()>p$, where a hyperparameter $p$ denotes the transform probability.
    By sampling different $(\alpha, \sigma)$ combinations at each iteration, the model is exposed to a wide range of geometric distortions, thereby enhancing robustness against implicit label noise.
        Besides, such stochastic sampling eliminates the burden of parameter tuning.

\subsection{\textbf{\textit{NSegment+}}: Incorporating scale-aware perturbation strategy into \textbf{\textit{NSegment}}}

This subsection presents our extension, \textbf{\textit{NSegment+}}, which further improves robustness by preventing harmful distortions on small segmentation regions.
    While \textbf{\textit{NSegment}} uniformly applies deformation to all label regions, \textbf{\textit{NSegment+}} includes a scale-aware suppression module that selectively disables deformation for small masks.
The procedure extends NSegment by introducing a per-segment filter: \vspace{-0.1cm}
\begin{enumerate}
    \item For each segment ${S}_{j}$, if its pixel area is smaller than a predefined threshold $\theta$, a local neighborhood around the segment is identified (See line 14 in Algorithm 1).
    \vspace{-0.1cm}
    \item Within this neighborhood, the corresponding regions in $d\mathcal{X}$ and $d\mathcal{Y}$ are set to zero, effectively preserving the original label shape for small regions. (See lines 2-6 in Algorithm 1 or Fig. \ref{fig2}-\textcircled{3} for details)
    \vspace{-0.1cm}
\end{enumerate}

This enables small segments to undergo deformation in our framework, preventing semantic misalignment and performance degradation, especially in datasets with high object scale variance. 
    Please refer to Algorithm 1 and Fig. \ref{fig2} together for a full understanding of our framework.

\begin{algorithm}[!ht]
\label{algo1}
\caption{Procedure of \textbf{\textit{NSegment+}}.} \small
\SetKwInOut{Initialization}{Parameters}
\KwIn{Input image $\mathcal{I}$ of size $w \times h$, Its corresponding segmentation labels $\mathcal{L}=\{{S}_j\}_{j=1}^{C}$, Set of $(\alpha, \sigma)$ combinations $\mathbf{\Omega} = \{(\alpha_k, \sigma_k)\}_{k=1}^{K}$, Transform probability $p$, small mask threshold $\theta$}
\SetKwProg{Fn}{function}{:}{\KwRet}
\Fn{\textnormal{{\textit{suppressSmallMask}}(${S}_j, d\mathcal{X}, d\mathcal{Y}, \alpha$)}}{
    Let $x_{\textit{min}}, y_{\textit{min}}, x_{\textit{max}}, y_{\textit{max}}$ denote the minimum and maximum values of the $x$ and $y$ coordinates among all pixel points in segment ${S}_j$\;
    $\epsilon \gets \lfloor \alpha/2 \rfloor$\;
    $x_{\textit{min}}, y_{\textit{min}} \gets {\textit{max}}(x_{\textit{min}}-\epsilon, 0), {\textit{max}}(y_{\textit{min}}-\epsilon, 0)$\;
    $x_{\textit{max}}, y_{\textit{max}} \gets {\textit{min}}(x_{\textit{max}}+\epsilon, w), {\textit{min}}(y_{\textit{max}}+\epsilon, h)$\;
    Set $d\mathcal{X}[x, y]$ and $d\mathcal{Y}[x, y]$ to $0$ for all pixels $(x, y)$ \textit{s.t.} $x_{\textit{min}} \le x \le x_{\textit{max}}$ and $y_{\textit{min}} \le y \le y_{\textit{max}}$\;
}
    \If{${\textit{rand}}() > p$} {
        \textbf{return} $\mathcal{L}$\tcp*{\small\textnormal{No update labels}}  
    }
    Initialize ${S}_j^{*}$ for each category index $j$ as zero-filled matrix of shape $w \times h$\;
    $\alpha, \sigma \gets {\textit{randChoice}}(\mathbf{\Omega})$\;
    $d\mathcal{X} \gets G_{\sigma} \circledast \phi (\alpha (2 \cdot {\textit{rand}}(w, h) - 1) )$\;
    $d\mathcal{Y} \gets G_{\sigma} \circledast \phi (\alpha (2 \cdot {\textit{rand}}(w, h) - 1) )$\;
    \For{$j \gets 1$ \textbf{to} $C$}{
        \If{${S}_j.\textit{area} \le \theta$} {
            {\textit{suppressSmallMask}}(${S}_j, d\mathcal{X}, d\mathcal{Y}, \alpha$)\;
        }
        Set ${S}_j^{*}[\varphi(x + d\mathcal{X}[x, y]), \varphi(y + d\mathcal{Y}[x, y])]$ to ${S}_j[x, y] \text{ for each pixel } (x, y) \text{ in segment } {S}_j$\;
    }
$\mathcal{L}^* \gets (S_j^{*})_{j=1}^{C}$\;
\textbf{return} $\mathcal{L}^*$\tcp*{\small\textnormal{Update labels}} 
\end{algorithm}

\vspace{-0.1cm}
\section{Experiments and Analysis}

\begin{table*}[t!]
\vspace{-0.0cm} 
\centering
\caption{{\label{tab:table-name} Impact of \textbf{\textit{NSegment}} and \textbf{\textit{NSegment+}} on training state-of-the-art segmentation models on remote sensing scenes}} \vspace{-0.2cm}
\resizebox{13cm}{!}{%
\begin{tabular}{@{}c|ccc|ccc|ccc@{}}
\toprule
Datasets &
  \multicolumn{3}{c|}{Vaihingen} &
  \multicolumn{3}{c|}{Potsdam} &
  \multicolumn{3}{c}{LoveDA} \\ \midrule
Models &
  Baseline &
  \textbf{\textit{NSegment}} &
  \textbf{\textit{ NSegment+}} &
  Baseline &
  \textbf{\textit{NSegment}} &
  \textbf{\textit{NSegment+}} &
  Baseline &
  \textbf{\textit{NSegment}} &
  \textbf{\textit{NSegment+}} \\ \midrule
DeepLab V3+ (ECCV 18) &
  \begin{tabular}[c]{@{}c@{}}77.53$\pm$0.30\\\textcolor{white}{  } \end{tabular} &
  \begin{tabular}[c]{@{}c@{}}\underline{78.26}$\pm$0.63\\(\textcolor{red}{+0.73}) \end{tabular} &
  \begin{tabular}[c]{@{}c@{}}\textbf{78.34}$\pm$0.26\\(\textcolor{red}{+0.81}) \end{tabular} &
  \begin{tabular}[c]{@{}c@{}}82.95$\pm$0.10\\\textcolor{white}{  } \end{tabular} &
  \begin{tabular}[c]{@{}c@{}}\underline{83.16}$\pm$0.11\\(\textcolor{red}{+0.21}) \end{tabular} &
  \begin{tabular}[c]{@{}c@{}}\textbf{83.20}$\pm$0.03\\(\textcolor{red}{+0.25}) \end{tabular} &
  \begin{tabular}[c]{@{}c@{}}43.29$\pm$0.77\\\textcolor{white}{  } \end{tabular} &
  \begin{tabular}[c]{@{}c@{}}\underline{43.36}$\pm$0.62\\(\textcolor{red}{+0.07}) \end{tabular} &
  \begin{tabular}[c]{@{}c@{}}\textbf{43.60}$\pm$0.17\\(\textcolor{red}{+0.31}) \end{tabular} \\ \midrule
ANN (ICCV 19) &
  \begin{tabular}[c]{@{}c@{}}79.75$\pm$0.07\\\textcolor{white}{  } \end{tabular} &
  \begin{tabular}[c]{@{}c@{}}\underline{80.44}$\pm$0.12\\(\textcolor{red}{+0.69}) \end{tabular} &
  \begin{tabular}[c]{@{}c@{}}\textbf{80.60}$\pm$0.16\\(\textcolor{red}{+0.85}) \end{tabular} &
  \begin{tabular}[c]{@{}c@{}}84.81$\pm$0.06\\\textcolor{white}{  } \end{tabular} &
  \begin{tabular}[c]{@{}c@{}}\underline{84.84}$\pm$0.03\\(\textcolor{red}{+0.03}) \end{tabular} &
  \begin{tabular}[c]{@{}c@{}}\textbf{84.88}$\pm$0.13\\(\textcolor{red}{+0.07}) \end{tabular} &
  \begin{tabular}[c]{@{}c@{}}45.93$\pm$0.69\\\textcolor{white}{  } \end{tabular} &
  \begin{tabular}[c]{@{}c@{}}\underline{47.22}$\pm$0.19\\(\textcolor{red}{+1.29}) \end{tabular} &
  \begin{tabular}[c]{@{}c@{}}\textbf{47.32}$\pm$0.58\\(\textcolor{red}{+1.39}) \end{tabular} \\ \midrule
DANet (CVPR 19) &
  \begin{tabular}[c]{@{}c@{}}79.59$\pm$0.61\\\textcolor{white}{  } \end{tabular} &
  \begin{tabular}[c]{@{}c@{}}\underline{79.71}$\pm$0.24\\(\textcolor{red}{+0.12}) \end{tabular} &
  \begin{tabular}[c]{@{}c@{}}\textbf{79.93}$\pm$0.40\\(\textcolor{red}{+0.34}) \end{tabular} &
  \begin{tabular}[c]{@{}c@{}}84.47$\pm$0.24\\\textcolor{white}{  } \end{tabular} &
  \begin{tabular}[c]{@{}c@{}}\underline{84.51}$\pm$0.23\\(\textcolor{red}{+0.04}) \end{tabular} &
  \begin{tabular}[c]{@{}c@{}}\textbf{84.67}$\pm$0.14\\(\textcolor{red}{+0.20}) \end{tabular} &
  \begin{tabular}[c]{@{}c@{}}43.35$\pm$0.78\\\textcolor{white}{  } \end{tabular} &
  \begin{tabular}[c]{@{}c@{}}\underline{43.82}$\pm$0.89\\(\textcolor{red}{+0.47}) \end{tabular} &
  \begin{tabular}[c]{@{}c@{}}\textbf{44.16}$\pm$0.61\\(\textcolor{red}{+0.81}) \end{tabular} \\ \midrule
APCNet (CVPR 19) &
  \begin{tabular}[c]{@{}c@{}}79.15$\pm$0.85\\\textcolor{white}{  } \end{tabular} &
  \begin{tabular}[c]{@{}c@{}}\underline{80.39}$\pm$0.30\\(\textcolor{red}{+1.24}) \end{tabular} &
  \begin{tabular}[c]{@{}c@{}}\textbf{80.49}$\pm$0.35\\(\textcolor{red}{+1.34}) \end{tabular} &
  \begin{tabular}[c]{@{}c@{}}84.73$\pm$0.13\\\textcolor{white}{  } \end{tabular} &
  \begin{tabular}[c]{@{}c@{}}\underline{84.74}$\pm$0.06\\(\textcolor{red}{+0.01}) \end{tabular} &
  \begin{tabular}[c]{@{}c@{}}\textbf{84.88}$\pm$0.01\\(\textcolor{red}{+0.15}) \end{tabular} &
  \begin{tabular}[c]{@{}c@{}}46.60$\pm$0.42\\\textcolor{white}{  } \end{tabular} &
  \begin{tabular}[c]{@{}c@{}}\underline{46.79}$\pm$1.06\\(\textcolor{red}{+0.19}) \end{tabular} &
  \begin{tabular}[c]{@{}c@{}}\textbf{47.25}$\pm$1.12\\(\textcolor{red}{+0.65}) \end{tabular} \\ \midrule
GCNet (TPAMI 20) &
  \begin{tabular}[c]{@{}c@{}}79.60$\pm$0.78\\\textcolor{white}{  } \end{tabular} &
  \begin{tabular}[c]{@{}c@{}}\textbf{80.55}$\pm$0.36\\(\textcolor{red}{+0.95}) \end{tabular} &
  \begin{tabular}[c]{@{}c@{}}\underline{80.38}$\pm$0.05\\(\textcolor{red}{+0.78}) \end{tabular} &
  \begin{tabular}[c]{@{}c@{}}84.90$\pm$0.12\\\textcolor{white}{  } \end{tabular} &
  \begin{tabular}[c]{@{}c@{}}\underline{84.91}$\pm$0.17\\(\textcolor{red}{+0.01}) \end{tabular} &
  \begin{tabular}[c]{@{}c@{}}\textbf{84.92}$\pm$0.12\\(\textcolor{red}{+0.02}) \end{tabular} &
  \begin{tabular}[c]{@{}c@{}}46.46$\pm$0.60\\\textcolor{white}{  } \end{tabular} &
  \begin{tabular}[c]{@{}c@{}}\underline{46.65}$\pm$0.63\\(\textcolor{red}{+0.19}) \end{tabular} &
  \begin{tabular}[c]{@{}c@{}}\textbf{46.70}$\pm$1.34\\(\textcolor{red}{+0.24}) \end{tabular} \\ \midrule
OCRNet (ECCV 20) &
  \begin{tabular}[c]{@{}c@{}}75.39$\pm$0.79\\\textcolor{white}{  } \end{tabular} &
  \begin{tabular}[c]{@{}c@{}}\underline{76.80}$\pm$0.42\\(\textcolor{red}{+1.41}) \end{tabular} &
  \begin{tabular}[c]{@{}c@{}}\textbf{77.68}$\pm$0.36\\(\textcolor{red}{+2.29}) \end{tabular} &
  \begin{tabular}[c]{@{}c@{}}82.62$\pm$0.11\\\textcolor{white}{  } \end{tabular} &
  \begin{tabular}[c]{@{}c@{}}\underline{82.63}$\pm$0.22\\(\textcolor{red}{+0.01}) \end{tabular} &
  \begin{tabular}[c]{@{}c@{}}\textbf{82.71}$\pm$0.11\\(\textcolor{red}{+0.09}) \end{tabular} &
  \begin{tabular}[c]{@{}c@{}}44.26$\pm$0.66\\\textcolor{white}{  } \end{tabular} &
  \begin{tabular}[c]{@{}c@{}}\underline{44.28}$\pm$0.04\\(\textcolor{red}{+0.02}) \end{tabular} &
  \begin{tabular}[c]{@{}c@{}}\textbf{44.61}$\pm$0.08\\(\textcolor{red}{+0.35}) \end{tabular} \\ \midrule
Mask2Former (CVPR 22) &
  \begin{tabular}[c]{@{}c@{}}77.47$\pm$0.22\\\textcolor{white}{  } \end{tabular} &
  \begin{tabular}[c]{@{}c@{}}\underline{77.85}$\pm$0.11\\(\textcolor{red}{+0.38}) \end{tabular} &
  \begin{tabular}[c]{@{}c@{}}\textbf{77.93}$\pm$0.09\\(\textcolor{red}{+0.46}) \end{tabular} &
  \begin{tabular}[c]{@{}c@{}}82.73$\pm$0.39\\\textcolor{white}{  } \end{tabular} &
  \begin{tabular}[c]{@{}c@{}}\underline{83.06}$\pm$0.07\\(\textcolor{red}{+0.33}) \end{tabular} &
  \begin{tabular}[c]{@{}c@{}}\textbf{83.20}$\pm$0.46\\(\textcolor{red}{+0.47}) \end{tabular} &
  \begin{tabular}[c]{@{}c@{}}48.50$\pm$0.68\\\textcolor{white}{  } \end{tabular} &
  \begin{tabular}[c]{@{}c@{}}\underline{48.82}$\pm$0.37\\(\textcolor{red}{+0.32}) \end{tabular} &
  \begin{tabular}[c]{@{}c@{}}\textbf{48.84}$\pm$0.40\\(\textcolor{red}{+0.34}) \end{tabular} \\ \midrule
DOCNet (GRSL 23) &
  \begin{tabular}[c]{@{}c@{}}80.80$\pm$0.32\\\textcolor{white}{  } \end{tabular} &
  \begin{tabular}[c]{@{}c@{}}\underline{81.41}$\pm$0.15\\(\textcolor{red}{+0.61}) \end{tabular} &
  \begin{tabular}[c]{@{}c@{}}\textbf{81.42}$\pm$0.06\\(\textcolor{red}{+0.62}) \end{tabular} &
  \begin{tabular}[c]{@{}c@{}}85.61$\pm$0.05\\\textcolor{white}{  } \end{tabular} &
  \begin{tabular}[c]{@{}c@{}}\underline{85.69}$\pm$0.05\\(\textcolor{red}{+0.08}) \end{tabular} &
  \begin{tabular}[c]{@{}c@{}}\textbf{85.70}$\pm$0.03\\(\textcolor{red}{+0.09}) \end{tabular} &
  \begin{tabular}[c]{@{}c@{}}50.92$\pm$0.50\\\textcolor{white}{  } \end{tabular} &
  \begin{tabular}[c]{@{}c@{}}\underline{51.12}$\pm$0.42\\(\textcolor{red}{+0.20}) \end{tabular} &
  \begin{tabular}[c]{@{}c@{}}\textbf{51.40}$\pm$0.45\\(\textcolor{red}{+0.48}) \end{tabular} \\ \midrule
CAT-Seg (CVPR 24) &
  \begin{tabular}[c]{@{}c@{}}71.59$\pm$0.20\\\textcolor{white}{  } \end{tabular} &
  \begin{tabular}[c]{@{}c@{}}\underline{71.62}$\pm$0.09\\(\textcolor{red}{+0.03}) \end{tabular} &
  \begin{tabular}[c]{@{}c@{}}\textbf{71.66}$\pm$0.10\\(\textcolor{red}{+0.07}) \end{tabular} &
  \begin{tabular}[c]{@{}c@{}}82.49$\pm$0.16\\\textcolor{white}{  } \end{tabular} &
  \begin{tabular}[c]{@{}c@{}}\underline{82.58}$\pm$0.04\\(\textcolor{red}{+0.09}) \end{tabular} &
  \begin{tabular}[c]{@{}c@{}}\textbf{82.61}$\pm$0.04\\(\textcolor{red}{+0.12}) \end{tabular} &
  \begin{tabular}[c]{@{}c@{}}46.89$\pm$0.58\\\textcolor{white}{  } \end{tabular} &
  \begin{tabular}[c]{@{}c@{}}\underline{47.24}$\pm$0.39\\(\textcolor{red}{+0.35}) \end{tabular} &
  \begin{tabular}[c]{@{}c@{}}\textbf{47.46}$\pm$0.28\\(\textcolor{red}{+0.57}) \end{tabular} \\ \midrule
Golden (CVPR 25) &
  \begin{tabular}[c]{@{}c@{}}70.67$\pm$0.34\\\textcolor{white}{  } \end{tabular} &
  \begin{tabular}[c]{@{}c@{}}\underline{71.34}$\pm$0.53\\(\textcolor{red}{+0.67}) \end{tabular} &
  \begin{tabular}[c]{@{}c@{}}\textbf{71.62}$\pm$0.50\\(\textcolor{red}{+0.95}) \end{tabular} &
  \begin{tabular}[c]{@{}c@{}}78.31$\pm$0.26\\\textcolor{white}{  } \end{tabular} &
  \begin{tabular}[c]{@{}c@{}}\underline{78.51}$\pm$0.11\\(\textcolor{red}{+0.20}) \end{tabular} &
  \begin{tabular}[c]{@{}c@{}}\textbf{78.76}$\pm$0.10\\(\textcolor{red}{+0.45}) \end{tabular} &
  \begin{tabular}[c]{@{}c@{}}37.56$\pm$1.91\\\textcolor{white}{  } \end{tabular} &
  \begin{tabular}[c]{@{}c@{}}\underline{39.16}$\pm$0.92\\(\textcolor{red}{+1.60}) \end{tabular} &
  \begin{tabular}[c]{@{}c@{}}\textbf{39.94}$\pm$0.99\\(\textcolor{red}{+2.38}) \end{tabular} \\ \midrule
LOGCAN++ (TGRS 25) &
  \begin{tabular}[c]{@{}c@{}}80.97$\pm$0.08\\\textcolor{white}{  } \end{tabular} &
  \begin{tabular}[c]{@{}c@{}}\underline{81.04}$\pm$0.14\\(\textcolor{red}{+0.07}) \end{tabular} &
  \begin{tabular}[c]{@{}c@{}}\textbf{81.09}$\pm$0.06\\(\textcolor{red}{+0.12}) \end{tabular} &
  \begin{tabular}[c]{@{}c@{}}86.07$\pm$0.06\\\textcolor{white}{  } \end{tabular} &
  \begin{tabular}[c]{@{}c@{}}\underline{86.11}$\pm$0.14\\(\textcolor{red}{+0.04}) \end{tabular} &
  \begin{tabular}[c]{@{}c@{}}\textbf{86.12}$\pm$0.01\\(\textcolor{red}{+0.05}) \end{tabular} &
  \begin{tabular}[c]{@{}c@{}}50.59$\pm$0.13\\\textcolor{white}{  } \end{tabular} &
  \begin{tabular}[c]{@{}c@{}}\underline{50.77}$\pm$0.34\\(\textcolor{red}{+0.18}) \end{tabular} &
  \begin{tabular}[c]{@{}c@{}}\textbf{51.23}$\pm$0.56\\(\textcolor{red}{+0.64}) \end{tabular} \\ \bottomrule
\multicolumn{10}{r}{Note: For each data-model combination, the highest result is highlighted in \textbf{bold} whereas the second-best is \underline{underlined} in this paper} \\
\end{tabular}
} \vspace{-0.1cm}
\label{tab1}
\end{table*}

\subsection{Experimental setup}

\subsubsection{1) Datasets and metric}
To evaluate the effectiveness and generalization ability of our proposed framework, we conduct extensive experiments on six diverse semantic segmentation benchmarks: Vaihingen \cite{vaihingen}, Potsdam \cite{potsdam}, LoveDA \cite{wang2021loveda}, Cityscapes \cite{cordts2016cityscapes}, Pascal VOC 2012 \cite{everingham2015pascal}, and COCO-Stuff 10K \cite{caesar2018coco}.
    These datasets encompass both remote sensing imagery (Vaihingen, Potsdam, LoveDA) and natural scenes (Cityscapes, Pascal VOC, COCO-Stuff), enabling a comprehensive performance assessment across various domains.
For all experiments, we use mean Intersection-over-Union (mIoU) as the evaluation metric. 
    To ensure statistical reliability, each experiment is conducted five times with different random seeds, and we report the mean and standard deviation of mIoU values across the five runs.

\subsubsection{2) Implementation details}
We implement all our models using the MMSegmentation framework based on PyTorch. 
    Our method is implemented as a custom transformation module and integrated into the standard data pre-processing pipeline. 
To ensure a fair comparison, we follow the conventional training and evaluation protocols specific to each dataset.
    We evaluate the effectiveness of our method on a wide range of state-of-the-art segmentation models. 
We automatically set the Gaussian kernel size for each $\sigma$ using the formula  $2 \cdot \left\lfloor 3 \sigma \right\rceil + 1$, which aligns with OpenCV's standard.
    The mask deformation suppression threshold $\theta$ was set to 1000 across all datasets. 
        The transform probability $p$ was 0.5 by default, but set to 1.0 in Tables \ref{tab2} and \ref{tab3}.
            Besides, we denote by $\mathbf{\Omega} = {(\alpha_k, \sigma_k)}_k$ the Cartesian product of the two sets ${1, 15, 30, 50, 100}$ and ${3, 5, 10}$ based on the grid search.
Data- and model-specific training configurations (learning rate, batch size, image pre-processing etc.) are documented in the supp. material. \vspace{-0.2cm}

\subsubsection{3) Assumptions of implicit label noise}
In contrast to typical studies that concentrate on obvious (or explicit) annotation mistakes, we make a different set of assumptions: (1) all datasets are free of explicit label errors, and (2) every piece of data is inherently subject to implicit label noise because true labels are unobservable and sometimes vague \cite{chen2021noise, wang2021loveda}.
    Therefore, we adopt them in their original form for both training and evaluation. 
Each dataset may inherently contain distinct and unquantifiable forms of implicit label noise. 
    Hence, the consistent outperformance of our method over the baseline models (i.e., trained without \textbf{\textit{NSegment+}}) across these heterogeneous datasets provides compelling evidence of its robustness and practical relevance in real-world scenarios, where perfectly clean annotations cannot be guaranteed.


\subsection{Ablation study}
We conduct comprehensive ablation studies to validate the effectiveness of each component in our \textbf{\textit{NSegment+}} framework. 
Specifically, we analyze (1) the benefit of diverse deformation strengths via stochastic Gaussian smoothing, (2) the impact of different deformation targets, and (3) the role of the small mask deformation suppression. For (1) and (2), we adopt PSPNet \cite{zhao2017pyramid} and conduct experiments on the Vaihingen \cite{vaihingen} dataset.

\begin{table}[t!]
\vspace{-0.2cm} 
\centering
\caption{{\label{tab:table-name} Impact of different $\alpha$-$\sigma$ combinations for fixed and stochastic label-level elastic deformations in \textbf{\textit{NSegment}}}} \vspace{-0.2cm}
\resizebox{6cm}{!}{%
\begin{tabular}{@{}c|cccccc|c@{}}
\toprule
\multirow{3.5}{*}{Baseline} &
  \multicolumn{6}{c|}{Fixed $\alpha$-$\sigma$} &
  \multicolumn{1}{c}{\multirow{3.5}{*}{\begin{tabular}[c]{@{}c@{}}Stochastic $\alpha$-$\sigma$ \\ sampling\end{tabular}}} \\ \cmidrule(lr){2-7}
                       & \multicolumn{1}{c|}{\diagbox[]{$\sigma$}{$\alpha$}}   & 1     & 5     & 30    & 50    & 100   & \multicolumn{1}{l}{}   \\ \midrule
\multirow{3}{*}{77.41} & \multicolumn{1}{c|}{3}  & 77.57 & 77.50 & 77.51 & 77.25 & 75.99 & \multirow{3}{*}{\begin{tabular}[c]{@{}c@{}}\textbf{77.75}\\(\textcolor{red}{+0.34})\end{tabular}} \\
                       & \multicolumn{1}{c|}{5}  & 77.30 & 77.70 & 77.41 & 77.24 & 77.54 &                        \\
                       & \multicolumn{1}{c|}{10} & 76.49 & 77.34 & 77.67 & 77.55 & 77.66 &                        \\ \bottomrule
\end{tabular}
} \vspace{-0.1cm}
\label{tab2}
\end{table}

\begin{table}[t!]
\vspace{-0.0cm} 
\centering
\caption{{\label{tab:table-name} Effect of image vs. label-only elastic deformations in \textbf{\textit{NSegment}} for semantic segmentation}} \vspace{-0.2cm}
\resizebox{6.5cm}{!}{%
\begin{tabular}{@{}l|cc|l@{}}
\toprule
 \multicolumn{1}{c|}{\multirow{2.5}{*}{\begin{tabular}[c]{@{}c@{}}For all cases,\\ we adopt stochastic $\alpha$-$\sigma$ sampling\end{tabular}}} & \multicolumn{2}{c|}{Elastic deformation} & \multicolumn{1}{c}{\multirow{2.5}{*}{\begin{tabular}[c]{@{}c@{}}Test \\mIoU\end{tabular}}} \\ \cmidrule(lr){2-3}
 & Image & Label &  \\ \midrule
Baseline & -& -& 77.41\\
+ (a) Normal (Identical image-label transform) & \checkmark & \checkmark &  67.58 (\textcolor{blue}{-9.83})\\
+ (b) Transform only for images & \checkmark  &   & 70.69 (\textcolor{blue}{-6.72})\\
+ (c) Transform only for labels &  & \checkmark  & \textbf{77.75} (\textcolor{red}{+0.34})\\ \bottomrule
\end{tabular}
} \vspace{-0.2cm}
\label{tab3}
\end{table}

\begin{table}[t!]
\vspace{-0.0cm} 
\centering
\caption{{\label{tab:table-name} Average performance improvements over the baseline without \textbf{\textit{NSegment}} and \textbf{\textit{NSegment+}}. \textbf{\textit{NSegment+}} consistently achieves larger gains, highlighting the benefit of scale-aware deformation suppression }} \vspace{-0.2cm}
\resizebox{7.2cm}{!}{%
\begin{tabular}{@{}c|c|cccccc@{}}
\toprule
 & \begin{tabular}[c]{@{}c@{}}Scale-aware\\ deformation \\ suppression\end{tabular} & Vaihingen & Potsdam & LoveDA & \begin{tabular}[c]{@{}c@{}}City- \\ scapes \end{tabular} & \begin{tabular}[c]{@{}c@{}}PASCAL \\ VOC \\2012\end{tabular} & \begin{tabular}[c]{@{}c@{}}COCO-\\Stuff \\ 10K\end{tabular} \\ \midrule
\textit{\textbf{NSegment}} &  & 0.63 & 0.10 & 0.44 & 0.41 & 0.57 & 0.19 \\ \midrule
\textit{\textbf{NSegment+}} & \checkmark & \begin{tabular}[c]{@{}c@{}}\textbf{0.78}\\(\textcolor{red}{+0.15})\end{tabular} & \begin{tabular}[c]{@{}c@{}}\textbf{0.18}\\(\textcolor{red}{+0.08})\end{tabular} & \begin{tabular}[c]{@{}c@{}}\textbf{0.74}\\(\textcolor{red}{+0.30})\end{tabular} & \begin{tabular}[c]{@{}c@{}}\textbf{0.83}\\(\textcolor{red}{+0.42})\end{tabular} & \begin{tabular}[c]{@{}c@{}}\textbf{0.91}\\(\textcolor{red}{+0.34})\end{tabular} & \begin{tabular}[c]{@{}c@{}}\textbf{0.44}\\(\textcolor{red}{+0.25})\end{tabular} \\ \bottomrule
\end{tabular}
} \vspace{-0.2cm}
\label{tab4}
\end{table}

\subsubsection{1) Benefit of stochastic Gaussian smoothing}

Table \ref{tab2} shows that stochastic sampling of deformation parameters $(\alpha, \sigma)$ outperforms fixed configurations across the board for label-only deformations.
While fixed settings can lead to marginal improvements over the baseline, the best choice of $(\alpha, \sigma)$ remains unclear.
    For example, $(\alpha=5, \sigma=5)$ works reasonably on Vaihingen but may not generalize to other datasets or models.
In contrast, our stochastic smoothing strategy randomly samples $(\alpha, \sigma)$ at each iteration from a wide parameter range $\mathbf{\Omega}$.
     It brings two major benefits: i) consistently improving performance by exposing the model to diverse label deformations (i.e., label-level regularization); ii) eliminating the need for manual parameter tuning, given that the parameter space $\mathbf{\Omega}$ is reasonably configured.

\begin{table*}[ht!]
\vspace{-0.3cm} 
\centering
\caption{{\label{tab:table-name} Impact of \textbf{\textit{NSegment}} and \textbf{\textit{NSegment+}} on training state-of-the-art segmentation models on natural scenes}} \vspace{-0.2cm}
\resizebox{13cm}{!}{%
\begin{tabular}{@{}c|ccc|ccc|ccc@{}}
\toprule
Datasets &
  \multicolumn{3}{c|}{Cityscapes} &
  \multicolumn{3}{c|}{PASCAL VOC 2012} &
  \multicolumn{3}{c}{COCO-Stuff 10K} \\ \midrule
Models &
  Baseline &
  \textbf{\textit{NSegment}} &
  \textbf{\textit{ NSegment+}} &
  Baseline &
  \textbf{\textit{NSegment}} &
  \textbf{\textit{NSegment+}} &
  Baseline &
  \textbf{\textit{NSegment}} &
  \textbf{\textit{NSegment+}} \\ \midrule
DeepLab V3+ (ECCV 18) &
  \begin{tabular}[c]{@{}c@{}}56.95$\pm$0.91\\\textcolor{white}{ } \end{tabular}&
  \begin{tabular}[c]{@{}c@{}}\underline{57.41}$\pm$0.70\\(\textcolor{red}{+0.46})\end{tabular} &
  \begin{tabular}[c]{@{}c@{}}\textbf{57.58}$\pm$1.08\\(\textcolor{red}{+0.63})\end{tabular} &
  \begin{tabular}[c]{@{}c@{}}61.64$\pm$0.53\\\textcolor{white}{ }\end{tabular} &
  \begin{tabular}[c]{@{}c@{}}\underline{61.80}$\pm$0.55\\(\textcolor{red}{+0.16})\end{tabular} &
  \begin{tabular}[c]{@{}c@{}}\textbf{61.84}$\pm$0.67\\(\textcolor{red}{+0.20})\end{tabular} &
  \begin{tabular}[c]{@{}c@{}}21.87$\pm$0.11\\\textcolor{white}{ }\end{tabular} &
  \begin{tabular}[c]{@{}c@{}}\underline{21.94}$\pm$0.31\\(\textcolor{red}{+0.07})\end{tabular} &
  \begin{tabular}[c]{@{}c@{}}\textbf{21.96}$\pm$0.49\\(\textcolor{red}{+0.09})\end{tabular} \\ \midrule
ANN (ICCV 19) &
  \begin{tabular}[c]{@{}c@{}}62.20$\pm$0.87\\\textcolor{white}{ }\end{tabular} &
  \begin{tabular}[c]{@{}c@{}}\underline{62.33}$\pm$0.64\\(\textcolor{red}{+0.13})\end{tabular} &
  \begin{tabular}[c]{@{}c@{}}\textbf{63.33}$\pm$1.09\\(\textcolor{red}{+1.13})\end{tabular} &
  \begin{tabular}[c]{@{}c@{}}67.87$\pm$0.65\\\textcolor{white}{ }\end{tabular} &
  \begin{tabular}[c]{@{}c@{}}\underline{68.53}$\pm$0.22\\(\textcolor{red}{+0.66})\end{tabular} &
  \begin{tabular}[c]{@{}c@{}}\textbf{68.66}$\pm$0.98\\(\textcolor{red}{+0.79})\end{tabular} &
  \begin{tabular}[c]{@{}c@{}}29.50$\pm$0.25\\\textcolor{white}{ }\end{tabular} &
  \begin{tabular}[c]{@{}c@{}}\underline{29.51}$\pm$0.25\\(\textcolor{red}{+0.01})\end{tabular} &
  \begin{tabular}[c]{@{}c@{}}\textbf{29.63}$\pm$0.17\\(\textcolor{red}{+0.13})\end{tabular} \\ \midrule
DANet (CVPR 19) &
  \begin{tabular}[c]{@{}c@{}}62.91$\pm$0.38\\\textcolor{white}{ }\end{tabular} &
  \begin{tabular}[c]{@{}c@{}}\underline{64.21}$\pm$0.06\\(\textcolor{red}{+1.30})\end{tabular} &
  \begin{tabular}[c]{@{}c@{}}\textbf{64.66}$\pm$0.96\\(\textcolor{red}{+1.75})\end{tabular} &
  \begin{tabular}[c]{@{}c@{}}63.52$\pm$1.47\\\textcolor{white}{ }\end{tabular} &
  \begin{tabular}[c]{@{}c@{}}\underline{63.94}$\pm$1.93\\(\textcolor{red}{+0.42})\end{tabular} &
  \begin{tabular}[c]{@{}c@{}}\textbf{64.36}$\pm$1.72\\(\textcolor{red}{+0.84})\end{tabular} &
  \begin{tabular}[c]{@{}c@{}}27.94$\pm$0.26\\\textcolor{white}{ }\end{tabular} &
  \begin{tabular}[c]{@{}c@{}}\underline{28.02}$\pm$0.62\\(\textcolor{red}{+0.08})\end{tabular} &
  \begin{tabular}[c]{@{}c@{}}\textbf{28.13}$\pm$0.42\\(\textcolor{red}{+0.19})\end{tabular} \\ \midrule
APCNet (CVPR 19) &
  \begin{tabular}[c]{@{}c@{}}61.97$\pm$1.17\\\textcolor{white}{ }\end{tabular} &
  \begin{tabular}[c]{@{}c@{}}\underline{62.71}$\pm$0.14\\(\textcolor{red}{+0.74})\end{tabular} &
  \begin{tabular}[c]{@{}c@{}}\textbf{63.24}$\pm$0.59\\(\textcolor{red}{+1.27})\end{tabular} &
  \begin{tabular}[c]{@{}c@{}}63.04$\pm$0.64\\\textcolor{white}{ }\end{tabular} &
  \begin{tabular}[c]{@{}c@{}}\underline{63.45}$\pm$1.32\\(\textcolor{red}{+0.41})\end{tabular} &
  \begin{tabular}[c]{@{}c@{}}\textbf{63.82}$\pm$0.36\\(\textcolor{red}{+0.78})\end{tabular} &
  \begin{tabular}[c]{@{}c@{}}27.52$\pm$0.44\\\textcolor{white}{ }\end{tabular} &
  \begin{tabular}[c]{@{}c@{}}\underline{27.86}$\pm$0.33\\(\textcolor{red}{+0.34})\end{tabular} &
  \begin{tabular}[c]{@{}c@{}}\textbf{27.94}$\pm$0.36\\(\textcolor{red}{+0.42})\end{tabular} \\ \midrule
GCNet (TPAMI 20) &
  \begin{tabular}[c]{@{}c@{}}62.44$\pm$0.34\\\textcolor{white}{ }\end{tabular} &
  \begin{tabular}[c]{@{}c@{}}\underline{62.66}$\pm$0.52\\(\textcolor{red}{+0.22})\end{tabular} &
  \begin{tabular}[c]{@{}c@{}}\textbf{62.71}$\pm$0.11\\(\textcolor{red}{+0.27})\end{tabular} &
  \begin{tabular}[c]{@{}c@{}}61.07$\pm$3.17\\\textcolor{white}{ }\end{tabular} &
  \begin{tabular}[c]{@{}c@{}}\underline{63.50}$\pm$1.17\\(\textcolor{red}{+2.43})\end{tabular} &
  \begin{tabular}[c]{@{}c@{}}\textbf{64.46}$\pm$2.74\\(\textcolor{red}{+3.39})\end{tabular} &
  \begin{tabular}[c]{@{}c@{}}23.86$\pm$0.54\\\textcolor{white}{ }\end{tabular} &
  \begin{tabular}[c]{@{}c@{}}\underline{24.32}$\pm$0.11\\(\textcolor{red}{+0.46})\end{tabular} &
  \begin{tabular}[c]{@{}c@{}}\textbf{24.45}$\pm$0.47\\(\textcolor{red}{+0.59})\end{tabular} \\ \midrule
OCRNet (ECCV 20) &
  \begin{tabular}[c]{@{}c@{}}55.94$\pm$1.71\\\textcolor{white}{ }\end{tabular} &
  \begin{tabular}[c]{@{}c@{}}\underline{56.11}$\pm$1.52\\(\textcolor{red}{+0.17})\end{tabular} &
  \begin{tabular}[c]{@{}c@{}}\textbf{56.51}$\pm$1.21\\(\textcolor{red}{+0.57})\end{tabular} &
  \begin{tabular}[c]{@{}c@{}}59.65$\pm$1.39\\\textcolor{white}{ }\end{tabular} &
  \begin{tabular}[c]{@{}c@{}}\underline{59.78}$\pm$1.25\\(\textcolor{red}{+0.13})\end{tabular} &
  \begin{tabular}[c]{@{}c@{}}\textbf{60.23}$\pm$0.86\\(\textcolor{red}{+0.58})\end{tabular} &
  \begin{tabular}[c]{@{}c@{}}22.23$\pm$0.51\\\textcolor{white}{ }\end{tabular} &
  \begin{tabular}[c]{@{}c@{}}\underline{22.75}$\pm$0.30\\(\textcolor{red}{+0.52})\end{tabular} &
  \begin{tabular}[c]{@{}c@{}}\textbf{22.84}$\pm$0.28\\(\textcolor{red}{+0.61})\end{tabular} \\ \midrule
SegFormer (NeurIPS 21) &
  \begin{tabular}[c]{@{}c@{}}62.69$\pm$0.14\\\textcolor{white}{ }\end{tabular} &
  \begin{tabular}[c]{@{}c@{}}\underline{62.71}$\pm$0.08\\(\textcolor{red}{+0.02})\end{tabular} &
  \begin{tabular}[c]{@{}c@{}}\textbf{62.84}$\pm$0.52\\(\textcolor{red}{+0.15})\end{tabular} &
  \begin{tabular}[c]{@{}c@{}}69.70$\pm$0.37\\\textcolor{white}{ }\end{tabular} &
  \begin{tabular}[c]{@{}c@{}}\underline{69.85}$\pm$0.35\\(\textcolor{red}{+0.15})\end{tabular} &
  \begin{tabular}[c]{@{}c@{}}\textbf{70.00}$\pm$0.42\\(\textcolor{red}{+0.30})\end{tabular} &
  \begin{tabular}[c]{@{}c@{}}29.07$\pm$0.46\\\textcolor{white}{ }\end{tabular} &
  \begin{tabular}[c]{@{}c@{}}\underline{29.16}$\pm$0.26\\(\textcolor{red}{+0.09})\end{tabular} &
  \begin{tabular}[c]{@{}c@{}}\textbf{29.37}$\pm$0.06\\(\textcolor{red}{+0.30})\end{tabular} \\ \midrule
Mask2Former (CVPR 22) &
  \begin{tabular}[c]{@{}c@{}}62.98$\pm$1.12\\\textcolor{white}{ }\end{tabular} &
  \begin{tabular}[c]{@{}c@{}}\underline{63.50}$\pm$0.36\\(\textcolor{red}{+0.52})\end{tabular} &
  \begin{tabular}[c]{@{}c@{}}\textbf{63.72}$\pm$0.54\\(\textcolor{red}{+0.74})\end{tabular} &
  \begin{tabular}[c]{@{}c@{}}71.02$\pm$0.70\\\textcolor{white}{ }\end{tabular} &
  \begin{tabular}[c]{@{}c@{}}\underline{71.14}$\pm$0.40\\(\textcolor{red}{+0.12})\end{tabular} &
  \begin{tabular}[c]{@{}c@{}}\textbf{71.68}$\pm$0.38\\(\textcolor{red}{+0.66})\end{tabular} &
  \begin{tabular}[c]{@{}c@{}}32.84$\pm$0.31\\\textcolor{white}{ }\end{tabular} &
  \begin{tabular}[c]{@{}c@{}}\underline{32.93}$\pm$0.35\\(\textcolor{red}{+0.09})\end{tabular} &
  \begin{tabular}[c]{@{}c@{}}\textbf{34.13}$\pm$0.32\\(\textcolor{red}{+1.29})\end{tabular} \\ \midrule
Golden (CVPR 25) &
  \begin{tabular}[c]{@{}c@{}}74.09$\pm$0.50\\\textcolor{white}{ }\end{tabular} &
  \begin{tabular}[c]{@{}c@{}}\underline{74.21}$\pm$0.65\\(\textcolor{red}{+0.12})\end{tabular} &
  \begin{tabular}[c]{@{}c@{}}\textbf{75.06}$\pm$0.55\\(\textcolor{red}{+0.97})\end{tabular} &
  \begin{tabular}[c]{@{}c@{}}41.33$\pm$1.07\\\textcolor{white}{ }\end{tabular} &
  \begin{tabular}[c]{@{}c@{}}\underline{41.95}$\pm$0.58\\(\textcolor{red}{+0.62})\end{tabular} &
  \begin{tabular}[c]{@{}c@{}}\textbf{42.02}$\pm$0.20\\(\textcolor{red}{+0.69})\end{tabular} &
  \begin{tabular}[c]{@{}c@{}}10.92$\pm$0.39\\\textcolor{white}{ }\end{tabular} &
  \begin{tabular}[c]{@{}c@{}}\underline{10.93}$\pm$0.22\\(\textcolor{red}{+0.01})\end{tabular} &
  \begin{tabular}[c]{@{}c@{}}\textbf{11.28}$\pm$0.22\\(\textcolor{red}{+0.36})\end{tabular} \\ \bottomrule
\end{tabular}
} \vspace{-0.0cm}
\label{tab5}
\end{table*}

\begin{table*}[t!]
\vspace{-0.1cm} 
\centering
\caption{{\label{tab:table-name} Impact of combining \textbf{\textit{NSegment+}} and existing approaches, including image-level augmentation and logit-level regularization strategies across six diverse semantic segmentation benchmarks}} \vspace{-0.2cm}
\resizebox{13.3cm}{!}{%
\begin{tabular}{@{}c|c|c|c|c|c|c|c|c|c@{}}
\toprule
Dataset & Method & \begin{tabular}[c]{@{}c@{}}No augmentation/\\regularization used\end{tabular} & \begin{tabular}[c]{@{}c@{}}Horizontal \\ Flipping (HF)\end{tabular} & \begin{tabular}[c]{@{}c@{}}Random \\ Resize (RR)\end{tabular} & \begin{tabular}[c]{@{}c@{}}Photometric \\ Distortion (PD)\end{tabular} & CutOut (CO) & CutMix (CM) & \begin{tabular}[c]{@{}c@{}}Random \\ Erasing (RE)\end{tabular} & \begin{tabular}[c]{@{}c@{}}Label \\ Smoothing (LS)\end{tabular} \\ \midrule
\multirow{3.5}{*}{Vaihingen} & Baseline & \begin{tabular}[c]{@{}c@{}}77.53$\pm$0.30\end{tabular} & \begin{tabular}[c]{@{}c@{}}78.62$\pm$0.19\end{tabular} & \begin{tabular}[c]{@{}c@{}}80.04$\pm$0.38\end{tabular} & \begin{tabular}[c]{@{}c@{}}77.33$\pm$0.35\end{tabular} & \begin{tabular}[c]{@{}c@{}}78.29$\pm$0.69\end{tabular} & \begin{tabular}[c]{@{}c@{}}77.54$\pm$0.60\end{tabular} & \begin{tabular}[c]{@{}c@{}}79.01$\pm$0.27\end{tabular} & \begin{tabular}[c]{@{}c@{}}77.97$\pm$0.28\end{tabular} \\ \cmidrule{2-10}
 & w/ \textbf{ours} & \begin{tabular}[c]{@{}c@{}}\textbf{78.34}$\pm$0.26\\ (\textcolor{red}{+0.81})\end{tabular} & \begin{tabular}[c]{@{}c@{}}\textbf{78.95}$\pm$0.23\\ (\textcolor{red}{+0.33})\end{tabular} & \begin{tabular}[c]{@{}c@{}}\textbf{80.38}$\pm$0.13\\ (\textcolor{red}{+0.34})\end{tabular} & \begin{tabular}[c]{@{}c@{}}\textbf{77.68}$\pm$0.32\\ (\textcolor{red}{+0.35})\end{tabular} & \begin{tabular}[c]{@{}c@{}}\textbf{78.80}$\pm$0.33\\ (\textcolor{red}{+0.51})\end{tabular} & \begin{tabular}[c]{@{}c@{}}\textbf{78.34}$\pm$0.27\\ (\textcolor{red}{+0.80})\end{tabular} & \begin{tabular}[c]{@{}c@{}}\textbf{79.06}$\pm$0.20\\ (\textcolor{red}{+0.05})\end{tabular} & \begin{tabular}[c]{@{}c@{}}\textbf{78.47}$\pm$0.14\\ (\textcolor{red}{+0.50})\end{tabular} \\ \midrule
\multirow{3.5}{*}{Potsdam} & Baseline & \begin{tabular}[c]{@{}c@{}}82.95$\pm$0.10\end{tabular} & \begin{tabular}[c]{@{}c@{}}84.00$\pm$0.12\end{tabular} & \begin{tabular}[c]{@{}c@{}}84.37$\pm$0.13\end{tabular} & \begin{tabular}[c]{@{}c@{}}82.76$\pm$0.13\end{tabular} & \begin{tabular}[c]{@{}c@{}}83.27$\pm$0.03\end{tabular} & \begin{tabular}[c]{@{}c@{}}82.99$\pm$0.05\end{tabular} & \begin{tabular}[c]{@{}c@{}}83.43$\pm$0.04\end{tabular} & \begin{tabular}[c]{@{}c@{}}82.84$\pm$0.13\end{tabular} \\ \cmidrule{2-10}
 & w/ \textbf{ours} & \begin{tabular}[c]{@{}c@{}}\textbf{83.20}$\pm$0.03\\ (\textcolor{red}{+0.25})\end{tabular} & \begin{tabular}[c]{@{}c@{}}\textbf{84.12}$\pm$0.07\\(\textcolor{red}{+0.12})\end{tabular} & \begin{tabular}[c]{@{}c@{}}\textbf{84.44}$\pm$0.17\\ (\textcolor{red}{+0.07})\end{tabular} & \begin{tabular}[c]{@{}c@{}}\textbf{83.00}$\pm$0.03\\ (\textcolor{red}{+0.24})\end{tabular} & \begin{tabular}[c]{@{}c@{}}\textbf{83.31}$\pm$0.14\\ (\textcolor{red}{+0.04})\end{tabular} & \begin{tabular}[c]{@{}c@{}}\textbf{83.08}$\pm$0.05\\ (\textcolor{red}{+0.09})\end{tabular} & \begin{tabular}[c]{@{}c@{}}\textbf{83.53}$\pm$0.01\\ (\textcolor{red}{+0.10})\end{tabular} & \begin{tabular}[c]{@{}c@{}}\textbf{83.03}$\pm$0.19\\ (\textcolor{red}{+0.19})\end{tabular} \\ \midrule
\multirow{3.5}{*}{LoveDA} & Baseline & \begin{tabular}[c]{@{}c@{}}43.29$\pm$0.77\end{tabular} & \begin{tabular}[c]{@{}c@{}}43.48$\pm$0.22\end{tabular} & \begin{tabular}[c]{@{}c@{}}40.73$\pm$1.05\end{tabular} & \begin{tabular}[c]{@{}c@{}}41.86$\pm$1.51\end{tabular} & \begin{tabular}[c]{@{}c@{}}43.40$\pm$0.23\end{tabular} & \begin{tabular}[c]{@{}c@{}}43.21$\pm$0.83\end{tabular} & \begin{tabular}[c]{@{}c@{}}41.84$\pm$0.50\end{tabular} & \begin{tabular}[c]{@{}c@{}}43.05$\pm$0.47\end{tabular} \\ \cmidrule{2-10}
 & w/ \textbf{ours} & \begin{tabular}[c]{@{}c@{}}\textbf{43.60}$\pm$0.17\\ (\textcolor{red}{+0.31})\end{tabular} & \begin{tabular}[c]{@{}c@{}}\textbf{43.92}$\pm$0.24\\ (\textcolor{red}{+0.44})\end{tabular} & \begin{tabular}[c]{@{}c@{}}\textbf{45.89}$\pm$0.40\\ (\textcolor{red}{+5.16})\end{tabular} & \begin{tabular}[c]{@{}c@{}}\textbf{43.10}$\pm$0.39\\ (\textcolor{red}{+1.24})\end{tabular} & \begin{tabular}[c]{@{}c@{}}\textbf{43.91}$\pm$0.49\\ (\textcolor{red}{+0.51})\end{tabular} & \begin{tabular}[c]{@{}c@{}}\textbf{44.18}$\pm$0.53\\ (\textcolor{red}{+0.97})\end{tabular} & \begin{tabular}[c]{@{}c@{}}\textbf{42.19}$\pm$0.43\\ (\textcolor{red}{+0.35})\end{tabular} & \begin{tabular}[c]{@{}c@{}}\textbf{43.64}$\pm$0.25\\ (\textcolor{red}{+0.59})\end{tabular} \\ \midrule
\multirow{3.5}{*}{Cityscapes} & Baseline & \begin{tabular}[c]{@{}c@{}}56.95$\pm$0.91\end{tabular} & \begin{tabular}[c]{@{}c@{}}56.62$\pm$1.24\end{tabular} & \begin{tabular}[c]{@{}c@{}}72.95$\pm$0.27\end{tabular} & \begin{tabular}[c]{@{}c@{}}54.10$\pm$0.49\end{tabular} & \begin{tabular}[c]{@{}c@{}}55.92$\pm$0.15\end{tabular} & \begin{tabular}[c]{@{}c@{}}56.28$\pm$0.96\end{tabular} & \begin{tabular}[c]{@{}c@{}}56.28$\pm$0.16\end{tabular} & \begin{tabular}[c]{@{}c@{}}55.64$\pm$0.42\end{tabular} \\ \cmidrule{2-10}
 & w/ \textbf{ours} & \begin{tabular}[c]{@{}c@{}}\textbf{57.58}$\pm$1.08\\ (\textcolor{red}{+0.63})\end{tabular} & \begin{tabular}[c]{@{}c@{}}\textbf{58.35}$\pm$0.47\\ (\textcolor{red}{+1.73})\end{tabular} & \begin{tabular}[c]{@{}c@{}}\textbf{73.13}$\pm$0.44\\ (\textcolor{red}{+0.18})\end{tabular} & \begin{tabular}[c]{@{}c@{}}\textbf{54.60}$\pm$0.39\\ (\textcolor{red}{+0.50})\end{tabular} & \begin{tabular}[c]{@{}c@{}}\textbf{57.15}$\pm$0.68\\ (\textcolor{red}{+1.23})\end{tabular} & \begin{tabular}[c]{@{}c@{}}\textbf{56.92}$\pm$0.50\\ (\textcolor{red}{+0.64})\end{tabular} & \begin{tabular}[c]{@{}c@{}}\textbf{56.43}$\pm$0.99\\ (\textcolor{red}{+0.15})\end{tabular} & \begin{tabular}[c]{@{}c@{}}\textbf{56.56}$\pm$0.83\\ (\textcolor{red}{+0.92})\end{tabular} \\ \midrule
\multirow{3.5}{*}{\begin{tabular}[c]{@{}c@{}}PASCAL VOC\\ 2012\end{tabular}} & Baseline & \begin{tabular}[c]{@{}c@{}}61.64$\pm$0.53\end{tabular} & \begin{tabular}[c]{@{}c@{}}61.80$\pm$0.59\end{tabular} & \begin{tabular}[c]{@{}c@{}}55.95$\pm$0.80\end{tabular} & \begin{tabular}[c]{@{}c@{}}61.30$\pm$0.23\end{tabular} & \begin{tabular}[c]{@{}c@{}}61.56$\pm$0.35\end{tabular} & \begin{tabular}[c]{@{}c@{}}61.59$\pm$0.86\end{tabular} & \begin{tabular}[c]{@{}c@{}}61.03$\pm$0.88\end{tabular} & \begin{tabular}[c]{@{}c@{}}62.55$\pm$0.72\end{tabular} \\ \cmidrule{2-10}
 & w/ \textbf{ours} & \begin{tabular}[c]{@{}c@{}}\textbf{61.84}$\pm$0.67\\ (\textcolor{red}{+0.20})\end{tabular} & \begin{tabular}[c]{@{}c@{}}\textbf{61.85}$\pm$0.39\\ (\textcolor{red}{+0.05})\end{tabular} & \begin{tabular}[c]{@{}c@{}}\textbf{56.94}$\pm$0.68\\ (\textcolor{red}{+0.99})\end{tabular} & \begin{tabular}[c]{@{}c@{}}\textbf{61.55}$\pm$0.51\\ (\textcolor{red}{+0.25})\end{tabular} & \begin{tabular}[c]{@{}c@{}}\textbf{61.61}$\pm$0.60\\ (\textcolor{red}{+0.05})\end{tabular} & \begin{tabular}[c]{@{}c@{}}\textbf{61.97}$\pm$0.36\\ (\textcolor{red}{+0.38})\end{tabular} & \begin{tabular}[c]{@{}c@{}}\textbf{61.37}$\pm$0.37\\ (\textcolor{red}{+0.34})\end{tabular} & \begin{tabular}[c]{@{}c@{}}\textbf{62.90}$\pm$0.44\\ (\textcolor{red}{+0.35})\end{tabular} \\ \midrule
\multirow{3.5}{*}{\begin{tabular}[c]{@{}c@{}}COCO-Stuff\\ 10K\end{tabular}} & Baseline & \begin{tabular}[c]{@{}c@{}}21.87$\pm$0.11\end{tabular} & \begin{tabular}[c]{@{}c@{}}21.67$\pm$0.48\end{tabular} & \begin{tabular}[c]{@{}c@{}}18.90$\pm$0.25\end{tabular} & \begin{tabular}[c]{@{}c@{}}21.37$\pm$0.09\end{tabular} & \begin{tabular}[c]{@{}c@{}}21.82$\pm$0.11\end{tabular} & \begin{tabular}[c]{@{}c@{}}21.81$\pm$0.28\end{tabular} & \begin{tabular}[c]{@{}c@{}}21.37$\pm$0.31\end{tabular} & \begin{tabular}[c]{@{}c@{}}21.73$\pm$0.33\end{tabular} \\ \cmidrule{2-10}
 & w/ \textbf{ours} & \begin{tabular}[c]{@{}c@{}}\textbf{21.96}$\pm$0.49\\ (\textcolor{red}{+0.09})\end{tabular} & \begin{tabular}[c]{@{}c@{}}\textbf{21.82}$\pm$0.36\\ (\textcolor{red}{+0.15})\end{tabular} & \begin{tabular}[c]{@{}c@{}}\textbf{19.08}$\pm$0.38\\ (\textcolor{red}{+0.18})\end{tabular} & \begin{tabular}[c]{@{}c@{}}\textbf{21.60}$\pm$0.19\\ (\textcolor{red}{+0.23})\end{tabular} & \begin{tabular}[c]{@{}c@{}}\textbf{21.85}$\pm$0.21\\ (\textcolor{red}{+0.03})\end{tabular} & \begin{tabular}[c]{@{}c@{}}\textbf{21.94}$\pm$0.32\\ (\textcolor{red}{+0.13})\end{tabular} & \begin{tabular}[c]{@{}c@{}}\textbf{21.55}$\pm$0.36\\ (\textcolor{red}{+0.18})\end{tabular} & \begin{tabular}[c]{@{}c@{}}\textbf{22.00}$\pm$0.13\\ (\textcolor{red}{+0.27})\end{tabular} \\ \bottomrule
\end{tabular}
} \vspace{-0.2cm}
\label{tab6}
\end{table*}

\subsubsection{2) Impact of different deformation targets}

Table \ref{tab3} reveals that applying identical elastic transforms to both image and label severely degrades performance (-9.83), highlighting the unrealistic appearance distortions for model training. 
    Similar phenomenon can be observed for image-only elastic transformation (-6.72).
Conversely, applying deformation only to the label not only avoids visual corruption but also improves mIoU by +0.34.

\subsubsection{3) Role of small-mask deformation suppression}
Table \ref{tab4} demonstrates the benefit of the scale-aware deformation suppression module, which selectively disables warping for small objects during label deformation.
    This table reports the average performance gain (in mIoU) over the baseline for each dataset, averaged across all segmentation models listed in Tables \ref{tab1} and \ref{tab5}.
For example, on the Cityscapes dataset, \textbf{\textit{NSegment}} improves mIoU by +0.41, whereas  \textbf{\textit{NSegment+}} achieves +0.83, resulting in an absolute delta of +0.42, due to the small mask suppression.
This pattern is consistent across all datasets.
    It clearly indicates that naively applying deformation to all regions—regardless of object scale—can degrade performance, particularly in cases of many small instances or over-fragmented masks.


\subsection{Impact of \textbf{\textit{NSegment}} and \textbf{\textit{NSegment+}} on training SOTA segmentation models}
We evaluate the impact of \textbf{\textit{NSegment}} and \textbf{\textit{NSegment+}} across a wide range of state-of-the-art semantic segmentation models and datasets.
    Specifically, we adopt 12 segmentation models, including DeepLab V3+ \cite{chen2018encoder}, ANN \cite{zhu2019asymmetric}, DANet \cite{fu2019dual}, APCNet \cite{he2019adaptive}, GCNet \cite{cao2020global}, OCRNet \cite{yuan2020object}, SegFormer \cite{xie2021segformer}, Mask2Former \cite{cheng2022masked}, DOCNet \cite{ma2023docnet}, CAT-Seg \cite{cho2024cat}, Golden \cite{yang2025golden}, and LOGCAN++ \cite{ma2025logcan++}.
        These models span both CNN-based and Transformer-based architectures from 2018 to 2025.
    Experiments are conducted on six diverse benchmarks, including remote sensing datasets (Vaihingen, Potsdam, LoveDA) and natural scene datasets (Cityscapes, VOC, COCO-Stuff), covering a broad spectrum of input modalities and annotation characteristics.
Tables \ref{tab1} and \ref{tab5} summarize the mIoU performance of each model with and without our proposed augmentation.
    \textbf{\textit{NSegment}} consistently improves performance over the vanilla baseline, while \textbf{\textit{NSegment+}} provides further gains by preserving the structural fidelity of small objects. 
For instance, on PASCAL VOC, \textbf{\textit{NSegment+}} improves GCNet by +3.39 mIoU over the baseline, compared to +2.43 with \textbf{\textit{NSegment}}.

\begin{figure*}[!t]
    \centering
    \includegraphics[width=15.5cm]{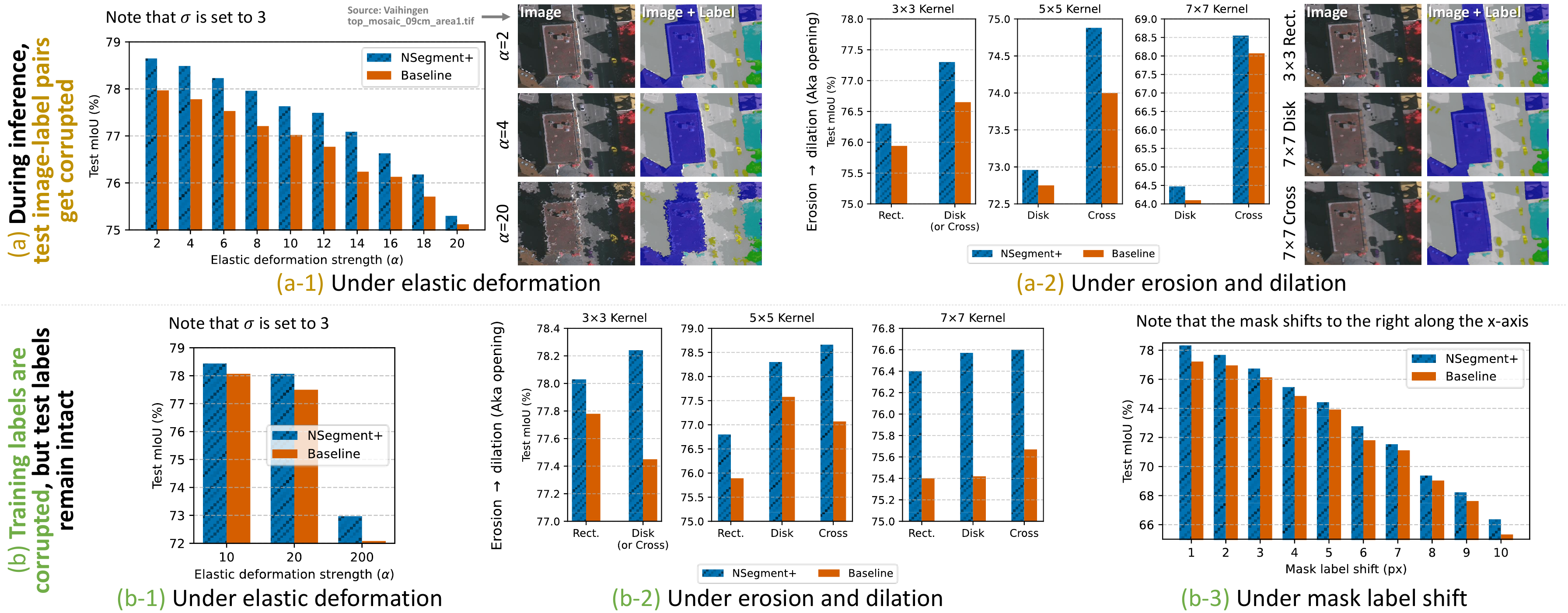} \vspace{-0.3cm}
    \caption{
    Robustness evaluation of \textbf{\textit{NSegment+}} to various types of implicit noise.
    Best viewed on a colored screen with zoom.}
    \label{fig3}
    \vspace{-0.3cm}
\end{figure*}

\subsection{Comparing \textbf{\textit{NSegment+}} with existing LNL approaches}

\begin{table}[t!]
\vspace{-0.0cm} 
\centering
\caption{{\label{tab:table-name} Comparing \textbf{\textit{NSegment+}} with existing LNL approaches designed for handling explicit label noise}} \vspace{-0.2cm}
\resizebox{6.5cm}{!}{%
\begin{tabular}{@{}l|l|l|l@{}}
\toprule
 & \multicolumn{1}{c|}{Vaihingen} & \multicolumn{1}{c|}{Cityscapes} & \multicolumn{1}{c}{GFLOPs} \\ \midrule
Baseline & 77.53$\pm$0.30 & 56.95$\pm$0.91 & \textcolor{white}{0}54.27 (1x)  \\ \midrule
Compensation Learning & 77.68$\pm$0.30 (\textcolor{red}{+0.15}) & 57.11$\pm$0.79 (\textcolor{red}{+0.16}) & \textcolor{white}{0}55.04 (1.01x) \\
L2B & 77.72$\pm$0.43 (\textcolor{red}{+0.19}) & 56.51$\pm$1.06 (\textcolor{blue}{-0.44}) & \textcolor{white}{0}54.27 (1x) \\
UCE & \underline{77.91}$\pm$0.12 (\textcolor{red}{+0.38}) & 57.01$\pm$1.57 (\textcolor{red}{+0.06}) & 542.69 (10x) \\
AIO2 & 77.73$\pm$0.27 (\textcolor{red}{+0.20}) & \underline{57.49}$\pm$1.14 (\textcolor{red}{+0.54}) & 108.54 (2x) \\ \midrule
\textit{\textbf{NSegment+}} & \textbf{78.34}$\pm$0.26 (\textcolor{red}{+0.81}) & \textbf{57.58}$\pm$1.08 (\textcolor{red}{+0.63}) & \textcolor{white}{0}54.27 (1x) \\ \bottomrule
\end{tabular}
} \vspace{-0.4cm}
\label{tab7}
\end{table}

Table \ref{tab7} benchmarks \textbf{\textit{NSegment+}} against state-of-the-art Learning from Noisy Labels (LNL) methods, including Compensation Learning \cite{kaiser2023compensation}, L2B \cite{zhou2024l2b}, UCE \cite{landgraf2024uncertainty}, and AIO2 \cite{liu2024aio2}, on Vaihingen and Cityscapes with DeepLab V3+.
    Compared to these LNL methods, often relying on uncertainty modeling, pseudo-label refinement, or auxiliary networks, \textbf{\textit{NSegment+}} is simpler and architecture-agnostic.
Besides, unlike UCE and AIO2, which require 10$\times$ and 2$\times$ more FLOPs, respectively, \textbf{\textit{NSegment+}} incurs no additional computational cost and achieves the best performance.
    These results highlight that implicit label noise, though often overlooked in prior LNL literature, deserves serious consideration.
Despite the superior standalone performance of \textbf{\textit{NSegment+}}, we believe that combining it with existing LNL strategies may offer complementary benefits, especially in scenarios with both implicit and explicit label noise.
   A thorough analysis of this is beyond the scope of this paper and is left for future work.



\subsection{Combining \textbf{\textit{NSegment+}} with existing augmentation or logit-level regularization strategies}

We examine whether \textbf{\textit{NSegment+}} complements existing augmentation techniques (e.g., CutOut \cite{devries2017improved}, CutMix \cite{yun2019cutmix}, Random Erasing \cite{zhong2020random}) and logit-level regularization (e.g., Label Smoothing \cite{szegedy2016rethinking}), across six benchmark datasets with DeepLab V3+.
    Table \ref{tab6} demonstrates that \textbf{\textit{NSegment+}} consistently improves mIoU when integrated with each strategy.
For example, on the LoveDA dataset, the combination of random resize and ours achieves a remarkable +5.16 mIoU improvement over using random resize alone. 
This is the highest boost among all variants tested on LoveDA.
    The strong synergy is likely because while random resize diversifies input scale at the image level, \textbf{\textit{NSegment+}} introduces implicit boundary noise at the label level—jointly pushing the model to learn scale-invariant and label-tolerant features under challenging rural-urban variations.

Besides, on PASCAL VOC, combining Label Smoothing \cite{szegedy2016rethinking} with \textbf{\textit{NSegment+}} yields the best performance, outperforming all other combinations (62.90 mIoU).
While Label Smoothing prevents overconfidence on hard-to-classify categories, \textbf{\textit{NSegment+}} encourages robustness to spatial imprecision—together forming a complementary regularization effect across both dimensions.


\vspace{-0.2cm}
\subsection{Robustness of \textbf{\textit{NSegment+}} to various implicit noise}

Fig. \ref{fig3} presents two complementary evaluations assessing the robustness of \textbf{\textit{NSegment+}} against various implicit noise, using DeepLab V3+ on Vaihingen.
Note that we preserve pixel-wise alignment between image and label for testing, enabling valid performance evaluation.
    In Fig. \ref{fig3}(a), we simulate scenarios in which both the input image and its corresponding label are synchronously perturbed through (a-1) elastic deformation and (a-2) morphological operations such as erosion and dilation. 
 These settings are designed to reflect test-time geometric distortions often encountered in real-world applications, such as sensor misalignment or preprocessing artifacts. 
    Across varying degrees of deformation, ours consistently yield higher mIoU than the baseline, demonstrating that training with label-specific deformations enhances robustness under test-time geometric distortions.

 In contrast, Fig. \ref{fig3}(b) evaluates a more challenging scenario in which only the training labels are perturbed to introduce synthetic but structurally plausible annotation noise—while input images remain clean.
    This setup mimics implicit label noise in various ranges arising from boundary ambiguity, inaccurate delineation, or annotator inconsistency. 
We avoid extreme or unrealistic cases such as class-flipping or mask removal, and instead focus on implicit noises caused by (b-1) elastic deformation, (b-2) erosion and dilation, and (b-3) pixel-level shifts. 
    Under these diverse perturbation patterns, \textbf{\textit{NSegment+}} significantly outperforms the baseline, indicating its effectiveness in learning noise-resilient representations from implicit noisy supervision.


\vspace{-0.2cm}
\subsection{Conclusions}
In this work, we proposed \textbf{\textit{NSegment+}}, a lightweight yet effective data augmentation strategy for semantic segmentation under implicit label noise.
    By decoupling image and label transformations—specifically, applying elastic deformations with stochastic Gaussian smoothing exclusively to segmentation labels—our method enables models to learn structure-aware and label-tolerant features.
To further enhance its robustness, we introduced a scale-aware suppression mechanism that mitigates semantic distortion for small objects, a critical factor in real-world datasets with high object scale variance.
    Extensive experiments across six diverse benchmarks—including remote sensing and natural scenes—demonstrate the broad applicability and consistent effectiveness of \textbf{\textit{NSegment+}}. 
Moreover, we showed that \textbf{\textit{NSegment+}} synergizes well with existing augmentation and regularization techniques, achieving additional gains without introducing computational overhead.

\vspace{-0.1cm}
\subsubsection{Discussions}
We note that elastic deformation may not always be the optimal choice for modeling mask ambiguities. 
    For instance, in building segmentation from aerial imagery, morphological operations like erosion may be the better option.
Furthermore, extending our proposed \textbf{\textit{NSegment+}} framework to tasks such as change detection or to other modalities, including medical imaging and synthetic datasets, represents a promising direction for future work.

\section{Acknowledgments} 
\begingroup
This work was enabled through collaboration with GIST-LIG Nex1, supported in part by the NRF grant (No. 2023R1A2C2006264), and benefited from high-performance GPU (A100) resources provided by HPC-AI Open Infrastructure via GIST SCENT. 
        A pilot version of our work appeared in \cite{kim2025nsegment}.
\endgroup







\section{Source Code}

For reproducibility and future work, we provide the implementation of \textbf{\textit{NSegment+}} using Python 3 and key libraries including NumPy, OpenCV2, and MMSegmentation as: \vspace{0.1cm}
\label{supalgo1}
\begin{lstlisting}[label=algo1,caption=Python implementation of our proposed algorithm named \textbf{\textit{NSegment+}}]
import cv2
import numpy as np

from mmseg.registry import TRANSFORMS


@TRANSFORMS.register_module()
class NoisySegmentPlus:
    def __init__(self, alpha_sigma_list=None, prob=0.5, 
                 area_thresh=1000):
        if not alpha_sigma_list:
            alpha_sigma_list = [(1, 3), (1, 5), (1, 10),
                                (15, 3), (15, 5), (15, 10),
                                (30, 3), (30, 5), (30, 10),
                                (50, 3), (50, 5), (50, 10),
                                (100, 3), (100, 5), (100, 10)]
        self.alpha_sigma_list = alpha_sigma_list
        self.prob = prob 
        self.area_thresh = area_thresh

    def __call__(self, results):
        if np.random.rand() > self.prob:
            return results
        segment = results['gt_seg_map']
        noisy_segment = self.transform(segment)
        results['gt_seg_map'] = noisy_segment
        return results

    def transform(self, segment, random_state=None):
        if random_state is None:
            random_state = np.random.RandomState(None)
        #+# [STEP 1] Generate random displacement fields
        shape = segment.shape[:2]
        dx = 2 * random_state.rand(*shape) - 1
        dy = 2 * random_state.rand(*shape) - 1

        #+# [STEP 2] Apply stochastic Gaussian smoothing
        _ = random_state.randint(0, len(self.alpha_sigma_list))
        alpha, sigma = self.alpha_sigma_list[_]
        dx, dy = alpha * dx, alpha * dy

        #+# [STEP 3] Scale-aware deformation suppression
        mask_ignore = np.zeros(shape, dtype=bool)
        unique_labels = np.unique(segment)
        for class_id in unique_labels:
            if class_id == -1:
                continue
            class_mask = (segment == class_id).astype(np.uint8)
            num_labels, labels = cv2.connectedComponents(
                class_mask
            )
            for comp_id in range(1, num_labels):
                comp_mask = (labels == comp_id).astype(np.uint8)
                area = np.sum(comp_mask)
                if area > area_thresh:
                    continue
                ys, xs = np.where(comp_mask)
                y1, x1 = ys.min(), xs.min()
                y2, x2 = ys.max(), xs.max()
                pad = alpha // 2
                y1_pad = max(y1-pad, 0)
                y2_pad = min(y2+pad, shape[0]-1)
                x1_pad = max(x1-pad, 0)
                x2_pad = min(x2+pad, shape[1]-1)
                mask_ignore[
                    y1_pad:y2_pad+1, x1_pad:x2_pad+1
                ] = True
        dx[mask_ignore], dy[mask_ignore] = 0, 0
        dx = cv2.GaussianBlur(dx, (0, 0), sigma)
        dy = cv2.GaussianBlur(dy, (0, 0), sigma)

        #+# [STEP 4] Label-specific (Mask) deformation
        x, y = np.meshgrid(
            np.arange(shape[1]), np.arange(shape[0])
        )
        map_x = np.clip(x+dx, 0, shape[1]-1).astype(np.float32)
        map_y = np.clip(y+dy, 0, shape[0]-1).astype(np.float32)
        noisy_segment = cv2.remap(segment, map_x, map_y, 
            interpolation=cv2.INTER_NEAREST
        )
        return noisy_segment
\end{lstlisting}

Note that, by removing lines 42 to 68, the above algorithm is reduced to the \textbf{\textit{NSegment}}.


\section{Descriptions of datasets used in this work}

\subsubsection{Vaihingen} consists of 33 true orthophoto (TOP) tiles with an average size of 2494×2064 and a spatial resolution of 9 cm. They were captured over a residential area in Vaihingen, Germany. 
    Following prior works, we adopt the standard split of 16 tiles for training and the remaining 17 tiles for testing.
We crop each tile into 512×512 with a stride of 256. 

\subsubsection{Potsdam} comprises 38 TOP tiles with a spatial resolution of 5 cm and a size of 6000×6000 pixels. 
    It covers a larger and more diverse urban area in Potsdam, Germany.
We follow the same data split protocol as in \cite{ma2025logcan++}, using 24 tiles for training and 14 for testing.
Both Vaihingen and Potsdam datasets provide pixel-level annotations for six semantic categories: \textit{impervious surfaces}, \textit{buildings}, \textit{low vegetation}, \textit{trees}, \textit{cars}, and \textit{clutter} (background). Since \textit{clutter} occupies only about 1\% of the data in both datasets, we exclude this category from experimentation.
    We crop each tile into 512×512 with a stride of 256. 

\subsubsection{LoveDA} is a large-scale land cover classification benchmark with 5987 annotated satellite images at a spatial resolution of 0.3 m. It includes both urban and rural scenes from multiple cities in China and covers seven semantic categories: \textit{background}, \textit{building}, \textit{road}, \textit{water}, \textit{barren}, \textit{forest}, and \textit{agriculture}.
    In contrast to the other two datasets, we retain the background class in LoveDA, following prior work.
We use 2522 images for training and 1669 for testing.
    Note that we utilize the official validation set for all testing and performance reporting, instead of relying on the online evaluation server.
Each image is randomly cropped into 512×512 for both training and testing.


\subsubsection{Cityscapes} is a high-resolution (2048×1024) urban scene dataset widely used for semantic segmentation, particularly in autonomous driving research. 
    It contains images from 50 European cities across varying seasons and weather conditions, split into 2975 training, 500 validation, and 1525 test images.
We follow the official 19-class (including \textit{road}, \textit{sidewalk}, \textit{building}, \textit{car}, and \textit{person}) setting \cite{cordts2016cityscapes} and use the validation set for evaluation. 
    Each image is randomly cropped into 512×1024 for training and testing.
    
\subsubsection{PASCAL VOC 2012} is a long-standing benchmark for visual recognition tasks, including object detection, classification, and semantic segmentation. 
    Due to the limited size of the original PASCAL VOC 2012 segmentation set, we adopt the augmented version of VOC 2012 by merging it with the Semantic Boundaries Dataset (SBD).
The augmented training set consists of 10582 natural images, including the original 1464 training images and 9118 additional samples with pixel-level annotations. 
    The validation set remains unchanged with 1449 images. 
Note that we utilize the official validation set for all testing and performance reporting, instead of relying on the online evaluation server.
Annotations cover 20 foreground object classes (e.g., \textit{person}, \textit{car}, \textit{dog}) and one background class, with fine-grained semantic labels provided at the pixel level.
    Each image is randomly cropped into 512×512 for training and testing.

\subsubsection{COCO-Stuff 10K} is a semantic segmentation benchmark that enriches the original COCO 2017 dataset by providing dense pixel-wise annotations for both \textit{thing} and \textit{stuff} classes. 
    While the original COCO dataset focuses primarily on instance-level annotations for countable objects (things), COCO-Stuff supplements this with detailed annotations for amorphous background regions (stuff), enabling holistic scene understanding.
The dataset consists of 10000 images of varying resolutions, split into 9000 training and 1000 test samples.
    Each pixel is labeled with one of 172 semantic classes, including 80 thing classes (e.g., \textit{person}, \textit{car}), 91 stuff classes (e.g., \textit{sky}, \textit{grass}, \textit{water}), and one unlabeled class.
Each image is randomly cropped into 512×512 for both training and testing.

\section{Training Configurations}

\subsection{For general experiments,}

To isolate the effect of \textbf{\textit{NSegment}} and \textbf{\textit{NSegment+}}  from variability introduced by data preprocessing, we controlled randomness in the cropping procedure by fixing the seed.
        Besides, no augmentation-related transformations were applied to either the baseline or \textbf{\textit{NSegment+}}, except in experiments explicitly targeting the interaction with data augmentation.
    The optimizer and learning rate scheduling strategies were kept as default, following the configurations specified in the publicly released implementations of the respective models.
Across all datasets, Golden \cite{yang2025golden} was trained for 120000 iterations, while all other models were trained for 80000 iterations, with batch size 4.

\subsection{For experiments relevant to comparing \textbf{\textit{NSegment+}} with existing LNL approaches,}


\begin{itemize}
    \item \textbf{T. Kaiser et al.}: We adopted their original configuration with \texttt{local\_compensation=True}, \texttt{loss\_balancing=0.01}, \texttt{non\_diagonal=True}, \texttt{symmetric=True}, and \texttt{top\_k=5}.
    
    \item \textbf{L2B}: Morphological bootstrapping was implemented using \texttt{cv2.erode} and \texttt{cv2.dilate}, with kernel sizes randomly sampled from \{7, 14, 21, 28, 35\} based on corresponding probabilities \{0.1, 0.2, 0.5, 0.7, 1.0\}. Area thresholding was applied using \texttt{cv2.connectedComponents} with a size threshold of 300. Random rotations were constrained to the angle range of (-0.01, 0.01) degrees.
    
    \item \textbf{UCE}: Each image was passed through the MC Dropout-activated network 10 times (\texttt{num\_samples=10}) to generate stochastic predictions.
    
    \item \textbf{AIO2}: We applied five types of mask-based augmentations—shift, erosion, dilation, rotation, and remove—prior to pseudo-label refinement. Area thresholding was performed with a minimum component size of 30. The rotation angle was randomly selected in the range (0.01, 1.0). Teacher and student have a same architecture.
\end{itemize}

\subsection{For experiments relevant to combining \textbf{\textit{NSegment+}} with existing image-level augmentation or logit-level regularization approaches,}

The following details were applied for each baseline:

\begin{itemize}
    \item \textbf{Horizontal Flipping (HF)}: This randomly mirrors the input along the horizontal axis (left-right). It was applied with a probability of 0.5.
    \item \textbf{Random Resize (RR)}: We applied random resizing with the aspect ratio preserved. The scale range was adjusted per dataset.
        For Cityscapes, each sample was resized within a range of 1.0 to 2.0. 
        For the remaining, each sample was resized within a scale ratio range of 0.5 to 2.0.
    \item \textbf{Photometric Distortion (PD)}: This introduces variations in image brightness, contrast, saturation, and hue. Unlike geometric transforms, this technique leaves the segmentation mask unchanged and operates only on the visual characteristics of the image. We used the default configuration provided by MMSegmentation.
    \item \textbf{CutOut (CO)}: It removes several rectangular patches from both the image and the corresponding label. These patches were randomly placed and filled with fixed values—black pixels in the image.
    The number of cutout holes varied between 5 and 10 per image, with patch sizes randomly chosen between 16×16 and 32×32 pixels, following \cite{devries2017improved}.
        This operation was applied with a probability 0.5 per image.
    \item \textbf{CutMix (CM)}: It replaces a rectangular region of the current sample with a patch from another randomly selected image in the batch.
        Both the image and segmentation label were modified in this manner. The size of the mixed region was determined by a randomly selected patch ratio between 20\% and 50\% of the total image area, following \cite{yun2019cutmix}. This form of CutMix was applied with a 50\% chance per training iteration.
    \item \textbf{Random Erasing (RE)}: It erases a single rectangular region in the image and its label. 
        CutOut fills the removed regions with a fixed black value, whereas Random Erasing replaces them with more diverse content, resulting in more natural-looking occlusions.
            The area to be erased was randomly selected with a size ratio between 5\% and 20\% of the entire image area, and an aspect ratio between 0.5 and 2.0, following \cite{zhong2020random}.
         The erased region was filled with either random pixel values, while simultaneously applying an ignore index (255) to the corresponding label regions.
            This operation was applied with a probability 0.5 per image.
    \item \textbf{Label Smoothing (LS)}: The semantic-level ground truth labels are softened by assigning a small probability mass (In this work, 0.1 was used) to non-target classes, thereby discouraging the model from becoming overconfident.
\end{itemize}

\end{document}